\documentclass{article}
\usepackage[accepted]{icml2022}

\usepackage[utf8]{inputenc} % allow utf-8 input
\usepackage[T1]{fontenc}    % use 8-bit T1 fonts
\usepackage{tabularx}
\usepackage{algorithm}
\usepackage{algorithmic}
\usepackage{capt-of}
\usepackage{url}
\usepackage{wrapfig}
\usepackage{microtype}
\usepackage{graphicx}
\usepackage{subfigure}
\usepackage{booktabs} % for professional tables
\usepackage[inline]{enumitem}
\usepackage{pifont}
\usepackage{multirow}
\usepackage{tikz}

\usepackage{xspace}
\usepackage{bbm}
\usepackage[space]{grffile}

\usepackage{authblk}
\usepackage{microtype}
\usepackage{graphicx}
\usepackage{subfigure}
\usepackage{booktabs} 
\usepackage{amsmath,amssymb, amsthm, bm}
\usepackage{color,soul}
\usepackage{hyperref}

\usetikzlibrary{tikzmark}
\newcommand{\mytikzmark}[1]{%
  \tikz[overlay,remember picture,baseline] \coordinate (#1) at (0,0) {};}
\newcommand{\highlight}[2]{%
  \draw[gray,line width=14pt,opacity=0.3]%
    ([yshift=4pt]#1) -- ([yshift=4pt]#2);%
}

\definecolor{citecolor}{rgb}{0.024,0.0784,0.627}

\hypersetup{
  colorlinks=true,      % false: boxed links; true: colored links
  linkcolor=citecolor,       % color of internal links
  citecolor=citecolor,    % color of links to bibliography
  filecolor=cyan,       % color of file links
  urlcolor=red          % color of external links
}

\usepackage[skins,listings]{tcolorbox}

\definecolor{dkgreen}{rgb}{0,0.6,0}
\definecolor{gray}{rgb}{0.5,0.5,0.5}
\definecolor{mauve}{rgb}{0.58,0,0.82}

\usepackage{hhline}
\usepackage{tikz}
\newcommand*\circled[1]{\tikz[baseline=(char.base)]{
            \node[shape=circle,draw,inner sep=0pt] (char) {#1};}}
            
\def\eg{{e.g.}}

\usepackage{textcomp}

\newenvironment{denseitemize}{
\begin{itemize}[topsep=2pt, partopsep=0pt, leftmargin=1.5em]
  \setlength{\itemsep}{2pt}
  \setlength{\parskip}{0pt}
  \setlength{\parsep}{0pt}
}{\end{itemize}}

\definecolor{babyblueeyes}{rgb}{0.63, 0.79, 0.95}

\def\name{FedScale\xspace}
\def\platform{\name Runtime\xspace}
\def\numdata{20\space}
\begin{document}
\twocolumn[
\icmltitle{\name: Benchmarking Model and System Performance of Federated Learning at Scale}

\begin{icmlauthorlist}
\icmlauthor{Fan Lai}{umich}
\icmlauthor{Yinwei Dai}{umich}
\icmlauthor{Sanjay S. Singapuram}{umich}
\icmlauthor{Jiachen Liu}{umich}
\icmlauthor{Xiangfeng Zhu}{umich,uw}
\icmlauthor{Harsha V. Madhyastha}{umich}
\icmlauthor{Mosharaf Chowdhury}{umich}
\end{icmlauthorlist}

\icmlaffiliation{umich}{Department of Computer Science, University of Michigan}
\icmlaffiliation{uw}{Department of Computer Science, University of Washington}
\icmlcorrespondingauthor{Fan Lai}{fanlai@umich.edu}
\icmlkeywords{Federated Learning, ICML}

\vskip 0.3in
]

\printAffiliationsAndNotice{} % otherwise use the standard text.

%\maketitle
\begin{abstract}

We present \name, a federated learning (FL) benchmarking suite with realistic datasets and a scalable runtime to enable reproducible FL research.
\name datasets encompass a wide range of critical FL tasks, ranging from image classification and object detection to language modeling and speech recognition.
Each dataset comes with a unified evaluation protocol using real-world data splits and evaluation metrics.
To reproduce realistic FL behavior, FedScale contains a scalable and extensible runtime.
It provides high-level APIs to implement FL algorithms, deploy them at scale across diverse hardware and software backends, and evaluate them at scale, all with minimal developer efforts.
We combine the two to perform systematic benchmarking experiments and highlight potential opportunities for heterogeneity-aware co-optimizations in FL.
\name is open-source and actively maintained by contributors from different institutions at \url{http://fedscale.ai}. 
We welcome feedback and contributions from the community.

\end{abstract}

\hypersetup{
  colorlinks=true,      % false: boxed links; true: colored links
  linkcolor=citecolor,       % color of internal links
  citecolor=citecolor,    % color of links to bibliography
  filecolor=cyan,       % color of file links
  urlcolor=citecolor          % color of external links
}

\section{Introduction}
\label{sec:intro}
\begin{figure}[t]%{r}{0.52\textwidth}
    \centering
    %\vspace{-.4cm}
    \includegraphics[width=\linewidth]{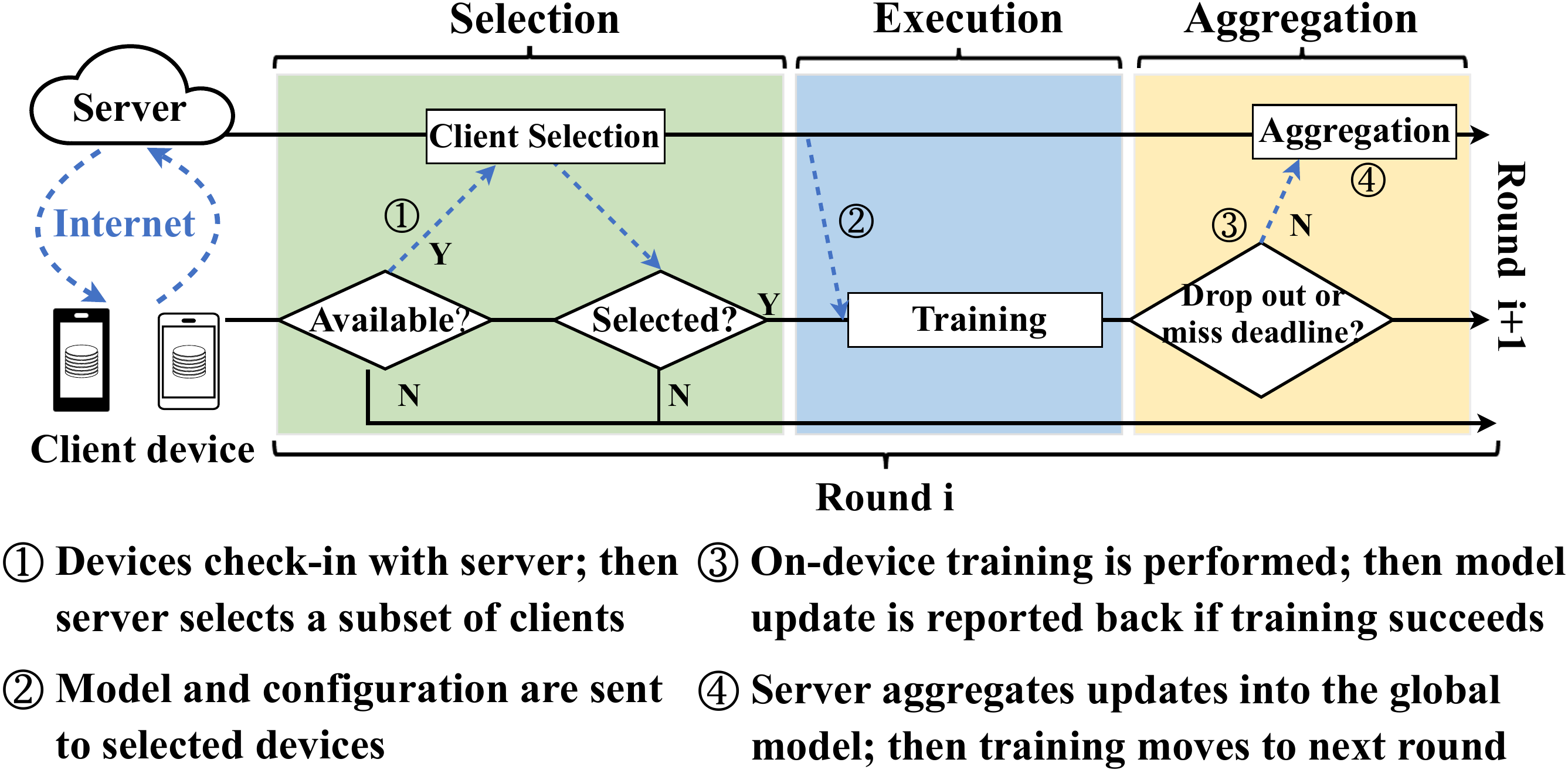}
    \caption{Standard FL protocol.}
    \label{fig:lifecycle}
    \vspace{-.6cm}
\end{figure}

Federated learning (FL) is an emerging machine learning (ML) setting where a logically centralized coordinator orchestrates many distributed 
clients (\eg, smartphones or laptops) to collaboratively train or evaluate a model \cite{federated-learning, fl-survey} (Figure~\ref{fig:lifecycle}). 
It enables model training and evaluation on end-user data, while circumventing high cost and privacy risks in gathering the raw data from clients, with applications across diverse ML tasks.
%: for example, NVIDIA applies FL to create medical imaging AI \cite{nvidia-fl}; Google runs federated training of NLP models in Google keyboard \cite{gg-oov, ggkeyboard}; Apple performs federated evaluation and tuning of automatic speech recognition models on end-user devices \cite{fl-apple}; IBM is deploying FL infrastructure to help detect financial misconducts \cite{fl-ibm}.

In the presence of heterogeneous execution speeds of client devices as well as non-IID data distributions,
existing efforts have focused on optimizing different aspects of FL: 
\begin{enumerate*}[label=({\arabic*})]
\item \emph{System efficiency}: 
reducing computation load (\eg, using smaller models~\cite{mobilenet}) or communication traffic (\eg, local SGD \cite{fedavg}) to achieve shorter round duration; 
\item \emph{Statistical efficiency}: 
designing data heterogeneity-aware algorithms (\eg, client clustering \cite{fl-clustering}) to obtain better training accuracy with fewer training rounds;  
\item \emph{Privacy and security}: 
developing reliable strategies (\eg, differentially private training \cite{dp-sgd}) to make FL more privacy-preserving and robust to potential attacks.
\end{enumerate*}

A comprehensive benchmark to evaluate an FL solution must investigate its behavior under the practical FL setting with 
\begin{enumerate*}[label=({\arabic*})]
    \item \emph{data heterogeneity} and
    \item \emph{device heterogeneity} under
    \item \emph{heterogeneous connectivity} and
    \item \emph{availability} conditions at
    \item \emph{multiple scales} on a
    \item \emph{broad variety of ML tasks}.
\end{enumerate*}
While the first two aspects are oft-mentioned in the literature~\cite{fed-prox}, realistic network connectivity and the availability of client devices can affect both types of heterogeneity (\eg, distribution drift~\cite{semi-sgd}), impairing model convergence.
Similarly, evaluation at a large scale can expose an algorithm's robustness, as practical FL deployment often runs across thousands of concurrent participants out of millions of clients~\cite{ggkeyboard}. 
Overlooking any one aspect can mislead FL evaluation (\S\ref{sec:related}).

Unfortunately, existing FL benchmarks often fall short across multiple dimensions (Table~\ref{table:comparison}).
First, they are limited in the versatility of data for various real-world FL applications. 
Indeed, even though they may have quite a few datasets and FL training tasks (\eg, LEAF~\cite{leaf-bench}), their datasets often contain synthetically generated partitions derived from conventional datasets (\eg, CIFAR) and do not represent realistic characteristics.
This is because these benchmarks are mostly borrowed from traditional ML benchmarks (\eg, MLPerf~\cite{mlperf}) or designed for simulated FL environments like TensorFlow Federated (TFF)~\cite{tff} or PySyft~\cite{pysyft}. 
Second, existing benchmarks often overlook system speed, connectivity, and availability of the clients (\eg, FedML~\cite{fedml} and Flower~\cite{flower}). 
This discourages FL efforts from considering system efficiency and leads to overly optimistic statistical performance (\S\ref{sec:related}).
Third, their datasets are primarily small-scale, because their experimental environments are unable to emulate large-scale FL deployments. 
While real FL often involves thousands of participants in each training round~\cite{fl-survey, ggkeyboard}, most existing benchmarking platforms can merely support the training of tens of participants per round.
Finally, most of them lack user-friendly APIs for automated integration, resulting in great engineering efforts for benchmarking at scale. 
We attach a detailed comparison of existing benchmarks against \name in Appendix~\ref{app:comparison}.

\newcommand{\incomplete}{$\bigcirc$}
\newcommand{\cmark}{\ding{52}}%
\newcommand{\xmark}{\ding{55}}%
\renewcommand{\arraystretch}{1.2}
\setlength{\tabcolsep}{2.15pt}

\begin{table}[t]
\small
\begin{center}
\begin{tabular*}{\linewidth}{cccccc}
\toprule  
Features          & LEAF & TFF & FedML & Flower  & \textbf{\name}\\
\midrule  
Heter. Client Dataset            &     \incomplete &       \xmark      &       \incomplete &   \incomplete     &   \cmark      \\
Heter. System Speed               &     \xmark      &       \xmark          &       \incomplete  &       \incomplete      &   \cmark      \\
Client Availability               &     \xmark      &       \xmark          &       \xmark  &       \xmark  &   \cmark      \\
Scalable Platform                 &     \xmark      &       \cmark         &       \incomplete & \cmark &  \cmark      \\  
Real FL Runtime                   &     \xmark      &       \xmark          &       \xmark  &       \xmark      &   \cmark      \\
Flexible APIs                 &     \xmark      &       \cmark          &       \cmark  &   \cmark      &   \cmark      \\  
\bottomrule 
\end{tabular*}
\end{center}
\caption{Comparing \name with existing FL benchmarks and libraries. \incomplete \ implies limited support.}
% \vspace{-.4cm}
\label{table:comparison}
\end{table}

% \begin{figure}[t]%{r}{0.4\textwidth}
%   \centering
%   %\vspace{-.25cm}
%   \includegraphics[scale=0.15]{figures/contri-cycle.pdf}
%   \caption{\name provides real FL data and an automated evaluation platform.}
%   \label{fig:overview}
%   %\vspace{-.6cm}
% \end{figure}

\paragraph{Contributions} 
We introduce an FL benchmark and accompanying runtime, \name, to enable comprehensive and standardized FL evaluations: 
\begin{denseitemize}
    \item To the best of our knowledge, \name presents the most comprehensive collection of FL datasets for evaluating different aspects of real FL deployments.
    It currently has \numdata realistic FL datasets with small, medium, and large scales for a wide variety of task categories, 
    such as image classification, object detection, word prediction, speech recognition, and reinforcement learning. 
    %We adopt domain-specific data splits based on the real client identification for each dataset. 
    To account for practical client behaviors, we include real-world measurements of mobile devices and associate each client with their  computation and communication speeds, as well as the availability status over time. 

    \item We present an automated evaluation platform, \platform, 
    to simplify and standardize FL evaluation in more realistic settings. 
    \platform provides a mobile backend to enable on-device FL evaluation and a cluster backend to benchmark various practical FL metrics (\eg, real client round duration) on GPUs/CPUs using real FL statistical and system datasets. The cluster backend can efficiently train thousands of clients in each round on a handful of GPUs. 
    \platform is also extensible, allowing easy deployment of new algorithms and ideas with flexible APIs. 

    \item We perform systematic experiments to show how \name facilitates comprehensive FL benchmarking and highlight the pressing need for co-optimizing system and statistical efficiency, especially in tackling system stragglers, accuracy bias, and device energy trade-offs. 
\end{denseitemize}
%
% !TeX root = ../flbench.tex
\section{Background}
\label{sec:related}

\begin{figure}
%\vspace{-.2cm}
\subfigure[Impact of system trace. \label{fig:impact-sys}]{\includegraphics[width=0.5\columnwidth]{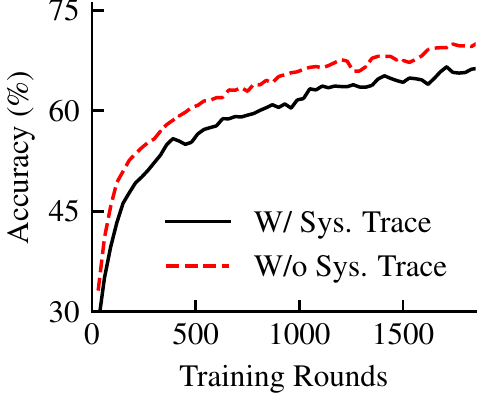}}\hfill
\subfigure[Impact of scale. \label{fig:impact-scale}]{\includegraphics[width=0.5\columnwidth]{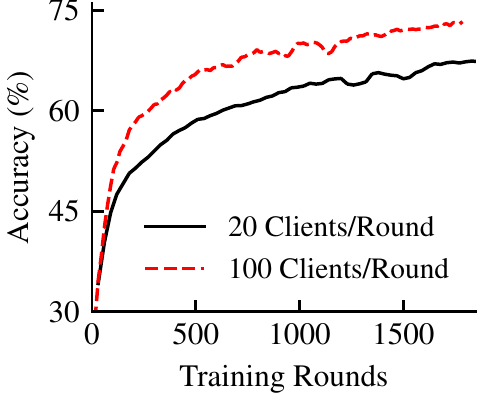}}
\caption{Existing benchmarks can be misleading. We train ShuffleNet on OpenImage classification (Detailed setup in Section~\ref{sec:eval}).}
% \vspace{-.5cm}
\end{figure}

\newcommand{\names}[1]{\emph{\small{#1}}}
\setlength{\tabcolsep}{2.5pt}

\begin{figure*}
% \small
\begin{minipage}[h]{\textwidth}
\begin{minipage}{0.73\textwidth}
\small
% \begin{center}
\centering
\begin{tabular}{cccrrc}
\toprule  
Category		& Name 		&Data Type 	& \#Clients	& \#Instances		& Example Task \\
\midrule  
\multirow{5}{*}{\textbf{CV}} 
 	%& \names{iNature}		&Image	&	2,295	&	193K		&	Classification	\\
 	%& \names{FEMNIST}		&Image	&	3,400	&	640K		&	Classification		\\
  	& \names{OpenImage}		&Image	&	13,771	&	1.3M		&	Classification, Object detection		\\	
  	& \names{Google Landmark}&Image	&	43,484	&	3.6M		&	Classification		\\
	& \names{Charades}		&Video	&	266		&	10K			&	Action recognition		\\
	& \names{VLOG}		&Video	&	4,900		&	9.6K	    &	Classification, Object detection		\\
	& \names{Waymo Motion}	&Video	&	496,358		&	32.5M			&	Motion prediction		\\	[.5ex]
\hline \\[-2.ex]
\multirow{5}{*}{\textbf{NLP}} 
	& \names{Europarl}		&Text	& 27,835		&	1.2M		&	Text translation	\\
	%& \names{Blog Corpus}	&Text	&   19,320		&	137M		&	Word prediction		\\
	%& \names{Stackoverflow}	&Text	&  342,477		&	135M		&	Word prediction, Classification	\\
	& \names{Reddit}		&Text	&  1,660,820	&	351M		&	Word prediction	\\
	%& \names{Amazon Review}	&Text	&  1,822,925	&	166M		&	Classification, Word prediction	\\
	%& \names{CoQA}			&Text 	& 7,189			&	114K		& 	Question Answering \\
	& \names{LibriTTS}		&Text	& 2,456			&	37K			&	Text to speech \\
	& \names{Google Speech} &Audio	& 2,618			&	105K		&	Speech recognition	\\
	& \names{Common Voice}	&Audio	& 12,976		&	1.1M		&	Speech recognition \\[.5ex]
\hline \\[-2.ex]
\multirow{3}{*}{\textbf{Misc ML}} 
	& \names{Taobao}		&Text	& 182,806		&	20.9M		&	Recommendation	\\
	%& \names{Didi}			&Text	& \todo{}		&	\todo{}		&	Reinforcement learning	\\
	& \names{Puffer Streaming}	&Text   &121,551		&  15.4M		& Sequence prediction \\
	& \names{Fox Go}	&Text	& 150,333		&	4.9M		&	Reinforcement learning	\\
\bottomrule 
\end{tabular}
\label{table:data-stats}
% \end{center}
\captionof{table}{Statistics of \emph{partial} \name datasets (the full list with more details is available in Appendix~\ref{app:dataset}). Currently, {\bf \name has \numdata real-world federated datasets}; each dataset is partitioned by its real client-data mapping, and we have removed sensitive information in these datasets.}
% \vspace{-.4cm}
\end{minipage}
\hspace{.45cm}
\begin{minipage}{0.24\textwidth}
% \vspace{-.2cm}
\begin{figure}[H]
	% \begin{minipage}{\columnwidth}
\centering
% \vspace{-.3cm}
\subfigure[\small{Data size.} \label{fig:data-size}]{\includegraphics[width=0.98\linewidth]{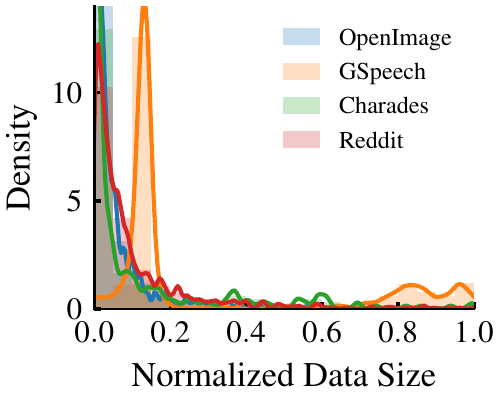}}
\subfigure[\small{Data distribution.} \label{fig:data-div}]{\includegraphics[width=0.98\linewidth]{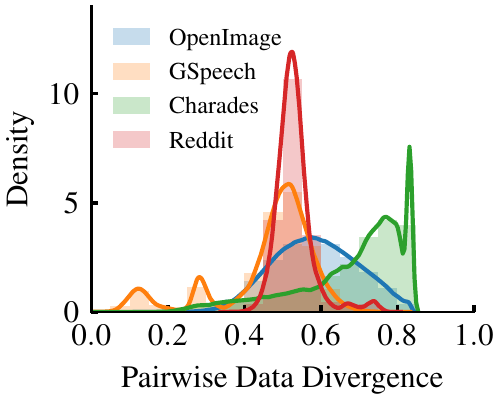}}
\vspace{-.3cm}
\captionof{figure}{Non-IID client data.}
\label{fig:data-heterogeneity}
\end{figure}
\end{minipage}
\end{minipage}
% \vspace{-.3cm}
\end{figure*}

\paragraph{Existing efforts toward practical FL}
	To tackle heterogeneous client data, FedProx~\cite{fed-prox}, FedYoGi~\cite{fed-yogi} and Scaffold~\cite{fed-scaffold} introduce adaptive client/server optimizations that use control variates to account for the `drift' in model updates. 
	Instead of training a single global model, 
	some efforts enforce guided client selection~\cite{kuiper-osdi}, train a mixture of models~\cite{fedbay, person-maml}, or cluster clients over training~\cite{fed-cluster}; 
	To tackle the scarce and heterogeneous device resource~\cite{sol}, FedAvg~\cite{fedavg} reduces communication cost by performing multiple local SGD steps, while some works compress the model update by filtering out or quantizing unimportant parameters~\cite{fetchsgd, model-compression}; 
	After realizing the privacy risk in FL~\cite{gradient-inf, backdoor-fl}, 
	DP-SGD~\cite{dpfl} enhances the privacy by employing differential privacy, 
	and DP-FTRL~\cite{dp-sgd} applies the tree aggregation to add noise to the sum of mini-batch gradients. 
These FL efforts often navigate accuracy-computation-privacy trade-offs. 
As such, a realistic FL setting is crucial for comprehensive evaluations. 

\paragraph{Existing FL benchmarks can be misleading}
Existing benchmarks often lack realistic client statistical and system behavior datasets and/or fail to reproduce large-scale FL deployments. 
As a result, they are not only insufficient for benchmarking diverse FL optimizations but can even mislead performance evaluations.
For example, 
\begin{enumerate*}[label=({\arabic*})]
\item As shown in Figure~\ref{fig:impact-sys}, statistical performance becomes worse when encountering realistic client behavior (\eg, training failures and availability dynamics), which indicates existing benchmarks that do not have systems traces can produce overly optimistic statistical performance;
\item FL training with hundreds of participants each round performs better than that with tens of participants (Figure~\ref{fig:impact-scale}). 
As such, existing benchmark platforms can under-report FL optimizations as they cannot support the practical FL scale with a large number of participants. 
\end{enumerate*}

% !TeX root = ../flbench.tex

\section{\name Dataset: Realistic FL Workloads}
\label{sec:data-trace}

We next introduce how we curate realistic datasets in \name to fulfill the desired properties of FL datasets.
%, such as the client statistical dataset and system traces.

\subsection{Client Statistical Dataset}
\label{subsec:training-data}

\name currently has \numdata realistic FL datasets (Table~\ref{table:data-stats}) across diverse practical FL scenarios and disciplines. For example, Puffer dataset~\cite{puffer} is from FL video streaming deployed to edge users over the Internet. %, spanning across a wide range of scales and task categories . 
The raw data of \name datasets are collected from different sources and stored in various formats.
We clean up the raw data, partition them into new FL datasets, streamline new datasets into consistent formats, 
and categorize them into different FL use cases.
Moreover, \name provides standardized APIs, a Python package, for the user to easily leverage these datasets (\eg, using different distributions of the same data or new datasets) in other frameworks.% (\eg, Tensorflow Federated).

\paragraph{Realistic data and partitions} 
We target realistic datasets with client information, and partition the raw dataset using the unique client identification. 
For example, OpenImage is a vision dataset collected by Flickr, wherein different mobile users upload their images to the cloud for public use. 
We use the \names{AuthorProfileUrl} attribute of the OpenImage data to map data instances to each client, 
whereby we extract the realistic distribution of the raw data. 
Following the practical FL deployments~\cite{ggkeyboard}, we assign the clients of each dataset into the training, validation and testing groups, to get its training, validation and testing set.
Here, we pick four real-world datasets -- video (Charades), audio (Google Speech), image (OpenImage), and text (Reddit) -- to illustrate practical FL characteristics. 
Each dataset consists of hundreds or up to millions of clients and millions of data points. 
Figure~\ref{fig:data-heterogeneity} reports the \emph{Probability Density Function} (PDF) of the data distribution, wherein we see a high statistical deviation (\eg, wide distribution of the density) across clients not only in the quantity of samples (Figure~\ref{fig:data-size}) but also in the data distribution (Figure~\ref{fig:data-div}).\footnote{We report the pairwise Jensen–Shannon distance of the label distribution between two clients.} 
We notice that realistic datasets mostly have unique Non-IID patterns, 
implying the impracticality of existing artificial FL partitions.

\paragraph{Different scales across diverse task categories}
To accommodate diverse scenarios in practical FL, \name includes small-, medium-, and large-scale datasets across a wide range of tasks, from hundreds to millions of clients. 
Some datasets can be applied in different tasks, as we enrich their use case by deriving different metadata from the same raw data.  
For example, the raw OpenImage dataset can be used for object detection, 
and we extract each object therein and generate a new dataset for image classification. 
Moreover, we provide APIs for the developer to customize their dataset (\eg, enforcing new data distribution or taking a subset of clients for evaluations with a smaller scale). 

\subsection{Client System Behavior Dataset}

\begin{figure}[t]
	\centering
	\subfigure[Compute capacity. \label{fig:computer-div}]{\includegraphics[width=0.5\linewidth]{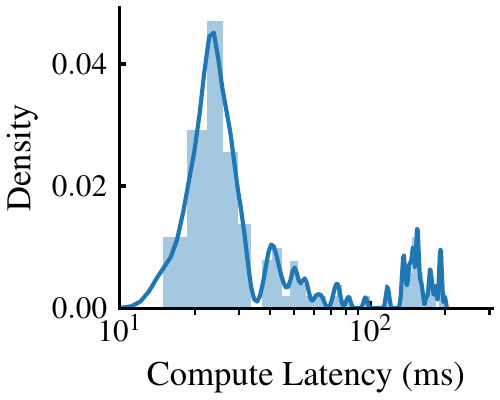}}\hfill
	\subfigure[Network capacity. \label{fig:bw-div}]{\includegraphics[width=0.5\linewidth]{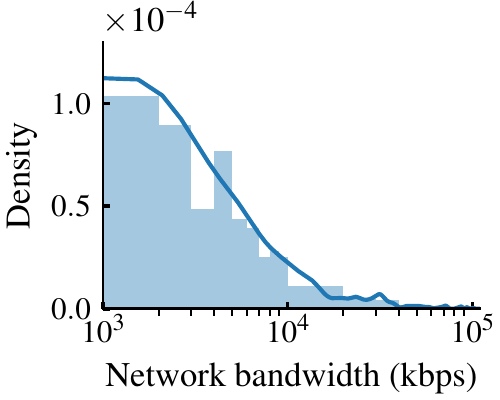}}
	\caption{Heterogeneous client system speed. }
	\label{fig:comp-bw-div}
	% \vspace{-.3cm}
\end{figure}

\begin{figure}[t]
\subfigure[Inter-device availability. \label{fig:diurnal-pattern}]{\includegraphics[width=0.5\columnwidth]{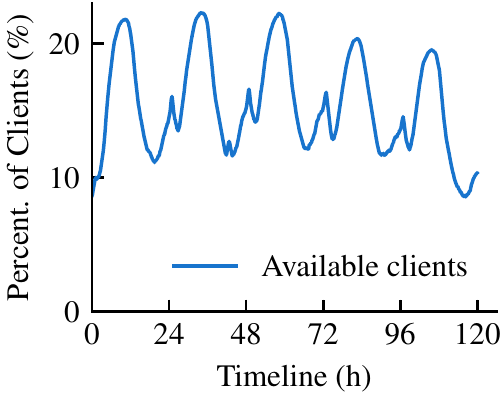}}\hfill
\subfigure[Intra-device availability. \label{fig:avail-dur}]{\includegraphics[width=0.5\columnwidth]{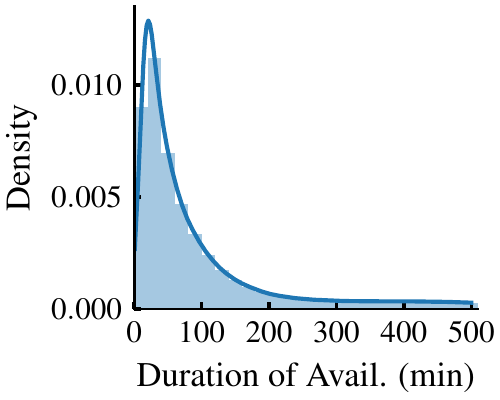}}
\caption{Client availability is dynamic.}
% \vspace{-.3cm}
\end{figure}

\paragraph{Client device system speed is heterogeneous} 
We formulate the system trace of different clients using \emph{AI Benchmark}~\cite{ai-bench} and 
\emph{MobiPerf Measurements}~\cite{mobiperf} on mobiles. 
\emph{AI Benchmark} provides the training and inference speed of diverse models (\eg, MobileNet) across a wide range of device models (\eg, Huawei P40 and Samsung Galaxy S20), while \emph{MobiPerf} has collected the available cloud-to-edge network throughput of over 100k world-wide mobile clients. 
As specified in real FL deployments~\cite{federated-learning, ggkeyboard}, 
we focus on mobile devices that have larger than 2GB RAM and connect with WiFi; Figure~\ref{fig:comp-bw-div} reports that their compute and network capacity can exhibit order-of-magnitude difference. 
As such, how to orchestrate scarce resources and mitigate stragglers are paramount.

\paragraph{Client device availability is dynamic}
We incorporate a large-scale user behavior dataset spanning across 136k users~\cite{client-avail} to emulate the behaviors of clients. 
It includes 180 million trace items of client devices (\eg, battery charge or screen lock) over a week. 
We follow the real FL setting, which considers the device in charging to be available \cite{federated-learning} and observe great dynamics in their availability: 
(i) the number of available clients reports diurnal variation (Figure~\ref{fig:diurnal-pattern}). 
This confirms the cyclic patterns in the client data, which can deteriorate the statistical performance of FL~\cite{semi-sgd}.
(ii) the duration of each available slot is not long-lasting (Figure~\ref{fig:avail-dur}). %, which is largely around tens of minutes. 
This highlights the need of handling failures (\eg, clients become offline) during training, as the round duration (also a few minutes) is comparable to that of each available slot. This, however, is largely overlooked in the literature.

% !TeX root = ../flbench.tex
\def\platform{FedScale Runtime\xspace}
\section{\platform: Evaluation Platform}
\label{sec:simulator}

Existing FL evaluation platforms are poor at reproducing practical, large-scale FL deployment scenarios. 
Worse, they often lack user-friendly APIs and require significant developer efforts to introduce new plugins. 
We introduce, \platform, an automated, extensible, and easily-deployable evaluation platform equipped with mobile and cluster backends, to simplify and standardize FL evaluation under realistic settings.

\newcommand{\tc}{\textdegree C\xspace}
\subsection{\platform: Mobile Backend}
\label{sec:mobile}

\platform deploys a mobile backend to enable on-device FL evaluation on smartphones. 
The first principle in building our mobile backend is to minimize any engineering effort for the developer (\eg, without reinventing their Python code) to benchmark FL on mobiles. 
To this end, \name mobile backend~\cite{swan-arxiv22} is built atop the Termux app~\cite{termux}, an Android terminal that supports Linux environment.

\begin{minipage}{\columnwidth}
\centering
\begin{lstlisting}
from fedscale.core.client import Client

class Mobile_Client(Client):
  def train(self,client_data,model,conf):
    for local_step in range(conf.local_steps):
      optimizer.zero_grad()
      ...
      loss.backward()
      optimizer.step()
    # Results will be sent to cloud aggregator via gRPC
    return gradient_update
\end{lstlisting}
  \captionof{figure}{Training on mobile client.}
\label{fig:mobile-client}
\end{minipage}

Figure~\ref{fig:mobile-client} shows a snippet of code running on \name mobile backend. By integrating with Termux, \platform allows the developer to run an unmodified version of Python script (\eg, PyTorch) built from source on the mobile device; the full-operator set (\eg, PyTorch modules) is available too. 
This speeds up the deployment cycle: FL models and algorithms that were prototyped on server GPUs/CPUs can also be deployed using \platform. 
We are currently implementing the Google Remote Procedure Call (gRPC) for distributed mobile devices to interface with \platform cloud server.

\begin{figure}[t]
\subfigure[Xiaomi Mi10 \label{fig:mobile-mi10}]{\includegraphics[width=0.5\linewidth]{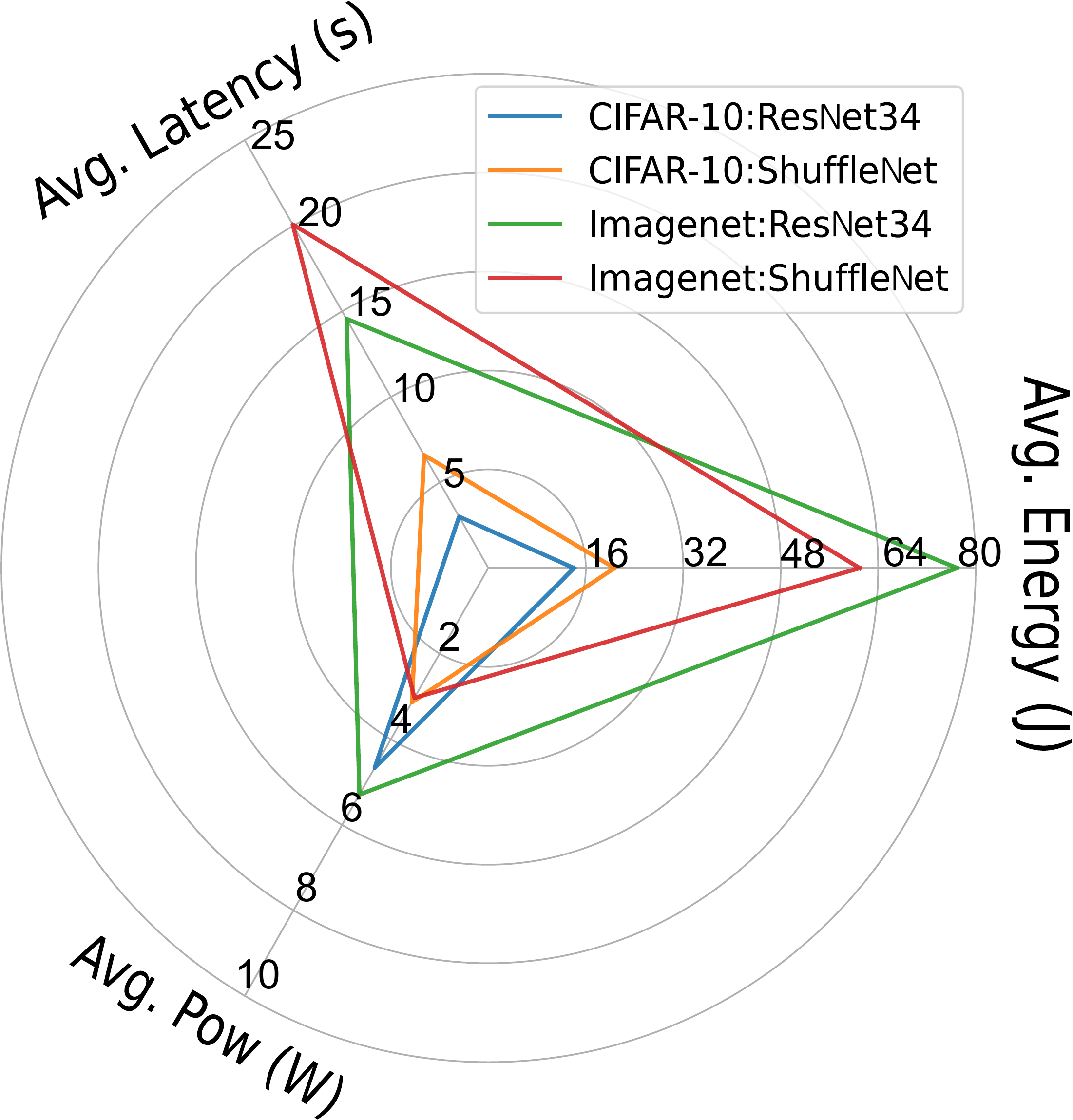}}\hfill
\subfigure[Samsung S10e \label{fig:mobile-s10e}]{\includegraphics[width=0.5\linewidth]{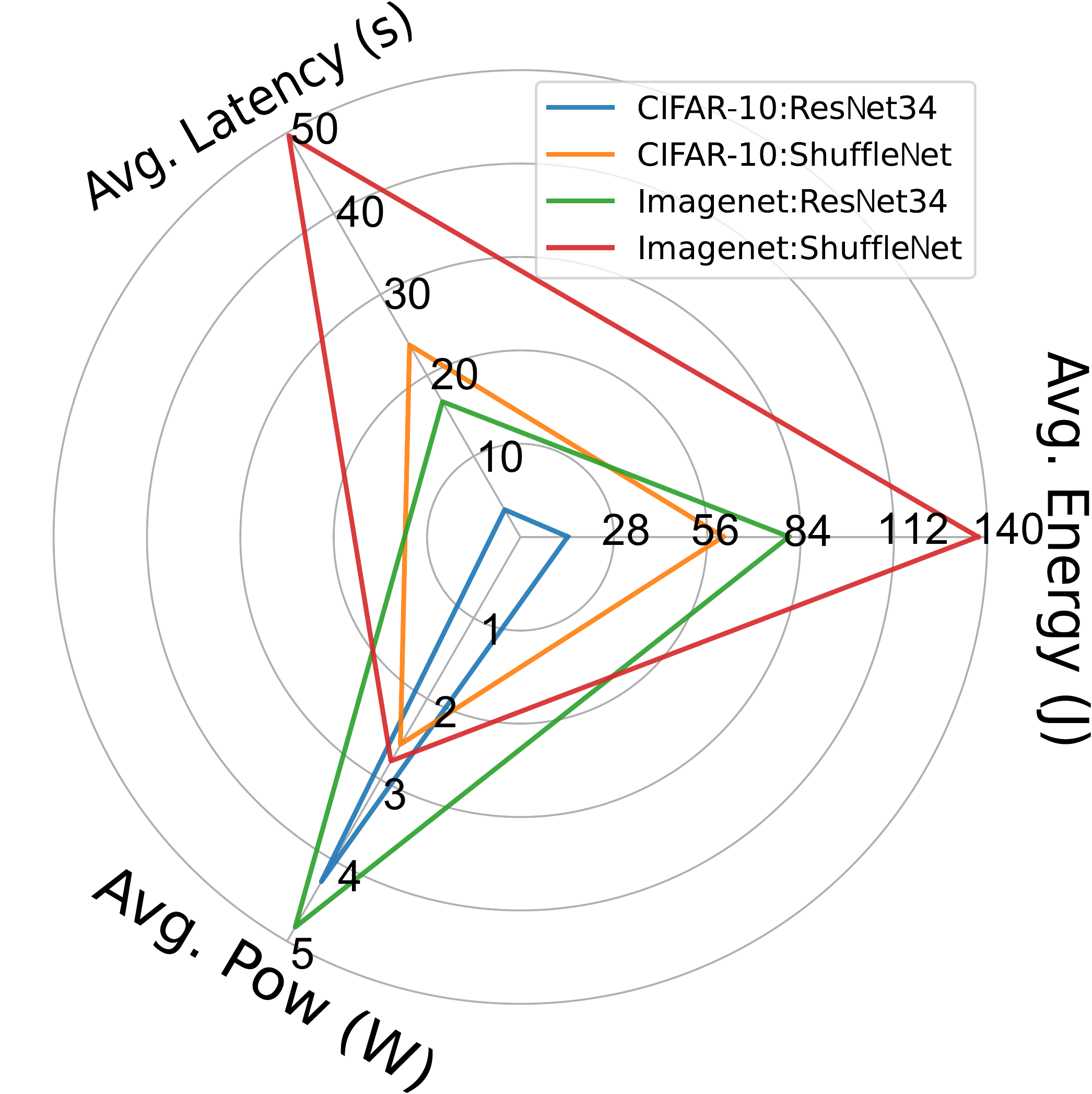}}
  \caption{\platform can benchmark the mobile runtime of power, energy and latency. We train Resnet34 and Shufflenet on ImageNet and CIFAR-10 on Xiaomi mi10 and Samsung S10e.}
\label{fig:mobile}
% \vspace{-.3cm}
\end{figure}

\paragraph{Benchmarking the mobile backend}
\name mobile backend enables developers to benchmark realistic FL training/testing performance on mobile phones. 
For example, Figure~\ref{fig:mobile} reports the performance metrics of training ShuffleNet and ResNet34 on one mini-batch (batch size 32), drawn from the ImageNet and CIFAR-10 datasets, on Xiaomi Mi10 and Samsung S10e Android devices. 
We benchmark the average training time.
We notice that ResNet34 runs at higher instantaneous power than ShuffleNet on both devices, but it requires less total energy to train since it takes shorter latency. 
ImageNet takes longer than CIFAR-10 per mini-batch, as the larger training image sizes lead to longer execution. 
The heterogeneity in computational capacity is evident as the Xiaomi Mi10 device outperforms the Samsung S10e device due to a more capable processor.
As such, we believe that \name mobile backend can facilitate future on-device FL optimizations (\eg, hardware-aware neural architecture search~\cite{amc}).

\subsection{\platform: Cluster Backend}
\label{sec:far-platform}
\platform provides an automated cluster backend that can support FL evaluations in real deployments and in-cluster simulations. 
In the \emph{deployment mode}, \platform acts as the cloud aggregator and orchestrates FL executions across real devices (\eg, laptops, mobiles, or even cloud servers). 
To enable cost-efficient FL benchmarking, \platform also includes a \emph{simulation mode} that performs FL training/testing on GPUs/CPUs, while providing various practical FL metrics by emulating realistic FL behaviors, such as computation/communication cost, latency and wall clock time. 
To the best of our knowledge, \platform is the first platform that enables FL benchmarking with practical FL runtime on GPUs/CPUs. 
More importantly, \platform can run the same code with little changes in both modes to minimize the migration overhead. 

Throughout the rest of the paper, we focus on the simulation mode as benchmarking is the primary focus of this paper.

\begin{figure}[t]%{r}{0.55\columnwidth}
\includegraphics[width=\columnwidth]{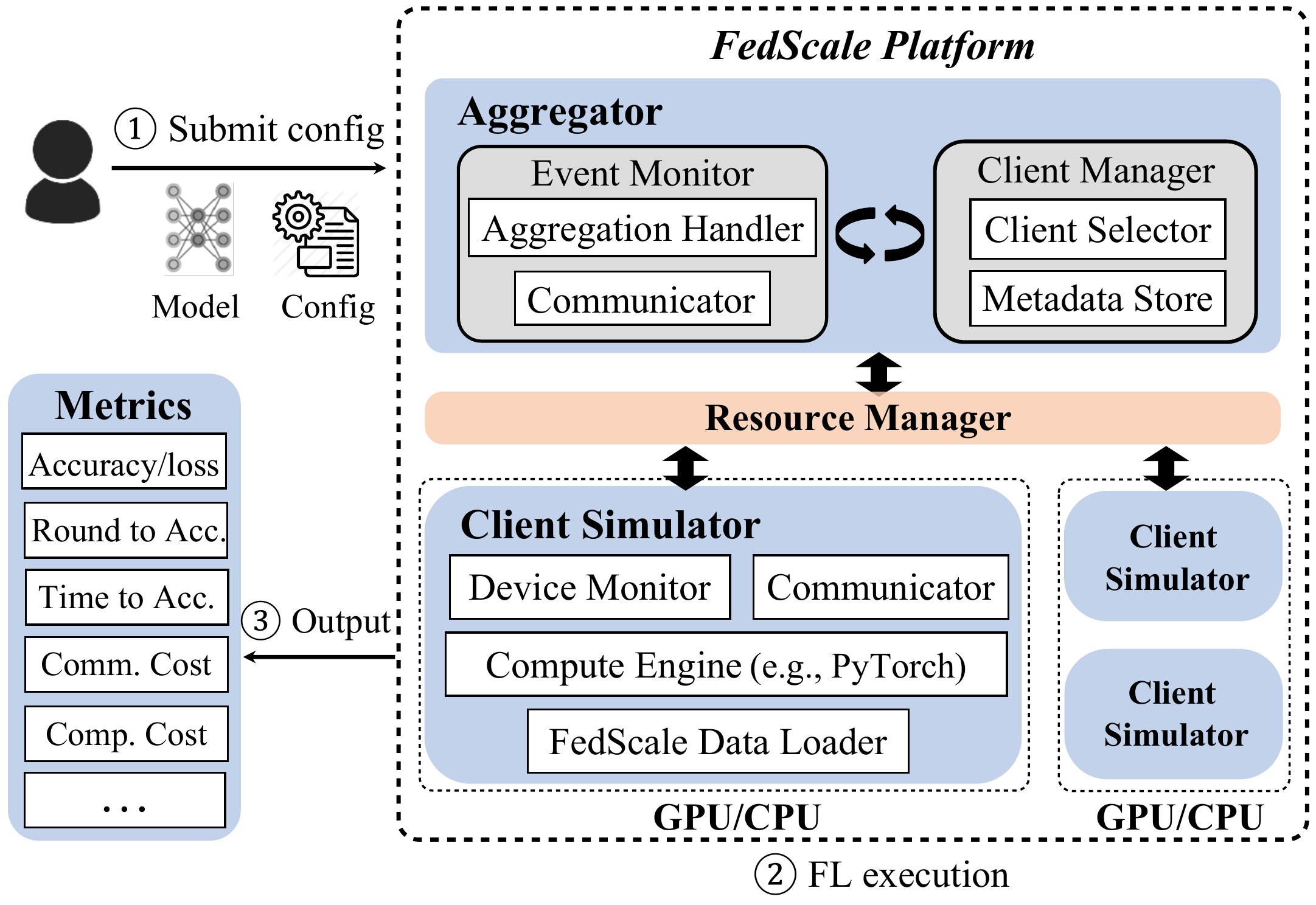}
\caption{\platform enables the developer to benchmark various FL efforts with practical FL data and metrics.}
\label{fig:arc}
% \vspace{-.4cm}
\end{figure}

\newcommand{\code}[1]{\small{\texttt{\color[rgb]{0.03,0,1}{#1}}}}
\begin{figure*}[t]
\begin{minipage}{0.58\textwidth}
\small
\begin{center}
\begin{tabular}{c|l|l}
\toprule  
Module &  API Name                            &  Example Use Case \\      
\midrule  
          %& \code{round\_initialization\_handler}(*args)     & Client clustering             \\
\textbf{Aggregator} & \code{round\_completion\_handler}()      & Adaptive/secure aggregation       \\  
\textbf{Simulator}  & \code{client\_completion\_handler}()   & Straggler mitigation      \\[.5ex]
          %& \code{serialize\_results}()     & Model compression   \\[.5ex]
%
\hline \\[-2.ex]
\textbf{Client}             & \code{select\_participants}()         & Client selection          \\
\textbf{Manager}            & \code{select\_model\_for\_client}()   & Adaptive model selection      \\[.5ex]
\hline \\[-2.ex]
\textbf{Client}   & \code{train}() & Local SGD/malicious attack      \\
\textbf{Simulator}    & \code{serialize\_results}()       & Model compression         \\
\bottomrule 
\end{tabular}
\end{center}
\captionof{table}{Some example APIs. \name provides APIs to deploy new plugins for various designs. We omit input arguments for brevity here.}
% \vspace{-.2cm}
\label{table:apis}
\end{minipage}
\hfill
\begin{minipage}{0.35\textwidth}
\centering
\begin{lstlisting}
from fedscale.core.client import Client

class Customized_Client(Client):
# Redefine training (e.g., for local SGD/gradient compression)

  def train(self,client_data,model,conf):
    # Code of plugin 
        ... 

    # Results will be serialized, and then sent to aggregator
    return training_result
\end{lstlisting}
  \captionof{figure}{Add plugins by inheritance.}
% \caption{Add plugins by inheritance.}
\label{fig:api}
\end{minipage}
% \vspace{-.3cm}

\end{figure*}

% \vspace{-.2cm}
\paragraph{Overview of the simulation mode}  
\platform consists of three primary components (Figure~\ref{fig:arc}):
\begin{denseitemize}
\item \emph{Aggregator Simulator}: 
It acts as the aggregator in practical FL, which selects participants, distributes execution profiles (\eg, model weight), 
and handles result (\eg, model updates) aggregation. 
In each round, its client manager uses the client behavior trace to monitor whether a client is available; then it selects the specified number of clients to participate that round. 
Once receiving new events, the event monitor activates the handler (\eg, aggregation handler to perform model aggregation), or 
the gRPC communicator to send/receive messages. 
The communicator records the size (cost) of every network traffic, and its runtime latency in FL wall-clock time ($\frac{traffic\_size}{client\_bandwidth}$).

\item \emph{Client Simulator}:
It works as the FL client. 
\name data loader loads the federated dataset of that client and feeds this data to the compute engine to run real training/testing. 
The computation latency is determined by ($\#\_processed\_sample \times latency\_per\_sample$), 
and the communicator handles network traffics and records the communication latency ($\frac{traffic\_size}{client\_bandwidth}$). 
%At the same time, the device monitor handles different function calls specified by the developer; 
The device monitor will terminate the simulation of a client if the current FL runtime exceeds his available slot (\eg, client drops out),   indicated by the availability trace. 

\item \emph{Resource Manager}: 
It orchestrates the available physical resource for evaluation to maximize the resource utilization. 
For example, when the number of participants/round exceeds the resource capacity (\eg, simulating thousands of clients on a few GPUs), the resource manager queues the overcommitted tasks of clients and schedules new client simulation from this queue once resource becomes available. Note that this queuing will not affect the simulated FL runtime, as this runtime is controlled by a global virtual clock, and the event monitor will manage events in the correct runtime order.
\end{denseitemize}

Note that capturing runtime performance (\eg, wall clock time) is rather slow and expensive in practical FL -- each mobile device takes several minutes to train a round -- but our simulator enables \emph{fast-forward} simulation, 
as training on CPUs/GPUs takes only a few seconds per round, while providing simulated runtime using realistic traces.

\paragraph{\platform enables automated FL simulation}  
\platform incorporates realistic FL traces, using the aforementioned trace by default or the developer-specified profile from the mobile backend, to automatically emulate the practical FL workflow:  
\circled{1} \emph{Task submission}: 
FL developers specify their configurations (\eg, model and dataset), 
which can be federated training or testing, and the resource manager 
 will initiate the aggregator and client simulator on available resource (GPU, CPU, other accelerators, or even smartphones); 
\circled{2} \emph{FL simulation}: 
Following the standardized FL lifecycle (Figure~\ref{fig:lifecycle}),  
in each training round, the aggregator inquires the client manager to select participants, 
whereby the resource manager distributes the client configuration to the available client simulators. 
After the completion of each client, the client simulator pushes the model update to the aggregator, 
which then performs the model aggregation. 
\circled{3} \emph{Metrics output}: 
During training, the developer can query the practical evaluation metrics on the fly. 
Figure~\ref{fig:arc} lists some popular metrics in \name.%; we attach more metrics and details in Appendix~\todo{}.

\paragraph{\platform is easily deployable and extensible with plugins}
\platform provides flexible APIs, which can accommodate with different execution backends (\eg, PyTorch) by design, 
for the developer to quickly benchmark new plugins.  
Table~\ref{table:apis} illustrates some example APIs that can facilitate diverse FL efforts, and  
Figure~\ref{fig:api} dictates an example showing how these APIs help to benchmark a new design of local client training with a few lines of code by inheriting the base Client module. 
Moreover, \platform can embrace new realistic (statistical client or system behavior) datasets with the built-in APIs. 
For example, the developer can import his own dataset of the client availability with the API (\texttt{load\_client\_availability}), and \platform will automatically enforce this trace during evaluations. 
We provide more examples and a comparison with other frameworks, in Appendix~\ref{app:plugin} to show the ease of evaluating various today's FL work in \name -- a few lines are all we need~\cite{kuiper-osdi, PyramidFL_MobiCom22}!

\begin{figure}[t]
% \begin{minipage}{.5\textwidth}
  \centering
  \includegraphics[width=.85\columnwidth]{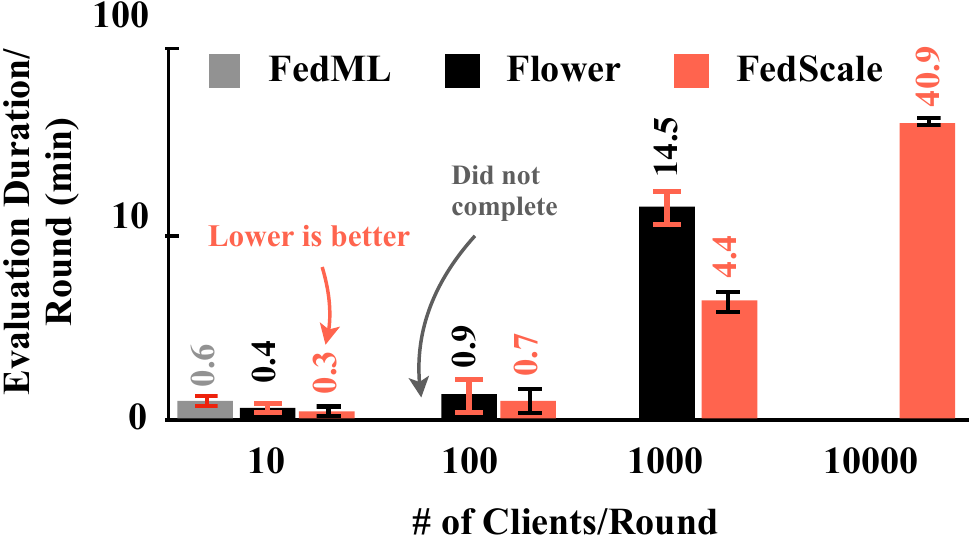}
  \caption{\platform can run thousands of clients/round on 10 GPUs where others fail. More results are in Appendix~\ref{app:comparison}. }
  \label{fig:scalability}
  % \vspace{-.3cm}
\end{figure}

\setlength{\tabcolsep}{8pt}
\begin{table*}
% \centering
\small
\centering
\begin{tabular}{c|c|c|c|c|c|c} 
\toprule
Task                 & Dataset       & Model               & IID     & FedAvg  & FedProx & FedYoGi  \\ 
  \midrule 
\hline
%   &      & MobileNet-V2      
% &  78.70\%  &57.90\%  &\textbf{70.60\%}&  63.70\%\\
 %    & 86.40\% & 78.50\% & 78.40\% & 76.30\%  \\ 
\cline{3-7} 
&      FEMNIST       & ResNet-18     & 86.40\% & \textbf{78.50\%} & 78.40\% & 76.30\%  \\ 
\cline{2-7} 
 &      & ShuffleNet-V2  & 81.37\%     & 70.27\%     & 69.54\%     & \textbf{74.04\%}      \\ 
\cline{3-7} 
\multirow{-3}{*}{Image Classification} & \multirow{-2}{*}{OpenImage} & MobileNet-V2  & 80.83\%     & 70.09\%     & 70.34\%     & \textbf{75.25\%}      \\ 
\hline
   Text Classification & Amazon Review & Logistic Regression & 66.10\% & \textbf{65.80\%} & 65.10\% & 65.30\%  \\ 
\hline
% \cline{2-7} 
     Language Modeling       & Reddit        & Albert    & 73.5 perplexity (ppl)   & 77.3 ppl   &  \textbf{76.6 ppl}   & 81.6 ppl    \\ 
\hline
Speech Recognition   & Google Speech & ResNet-34           & 72.58\% & \textbf{63.37\%}     & 63.25\%     & 62.67\%      \\
\bottomrule
\end{tabular}
 \caption{Benchmarking of different FL algorithms across realistic FL datasets. We report the mean test accuracy over 5 runs. }
 \label{table:eval}
% \vspace{-.2cm}
\end{table*}
\begin{figure*}[t]
\centering
  % \begin{minipage}{.9\linewidth}
   \centering
    \subfigure[Convergence on Google speech. \label{fig:rta-speech}]{\includegraphics[width=0.3\linewidth]{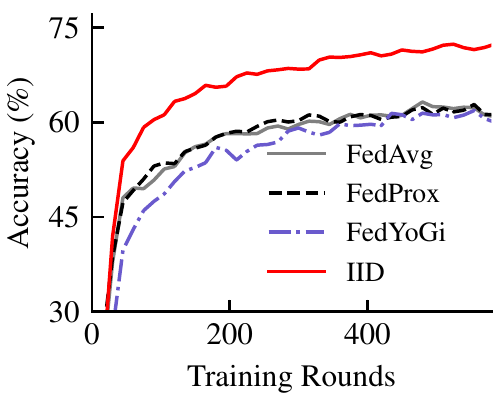}}\hfill
    \subfigure[Convergence on OpenImage. \label{fig:rta-image}]{\includegraphics[width=0.3\linewidth]{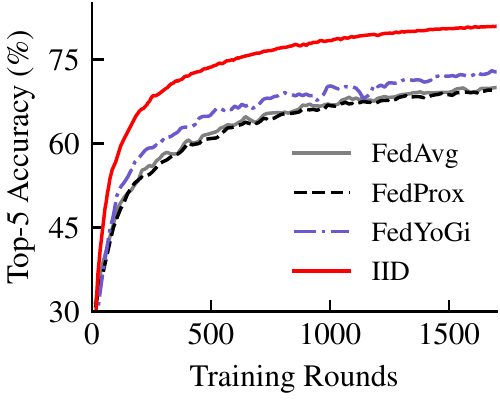}} \hfill
    \subfigure[Final model performance. \label{fig:impact-n}]{\includegraphics[width=0.3\linewidth]{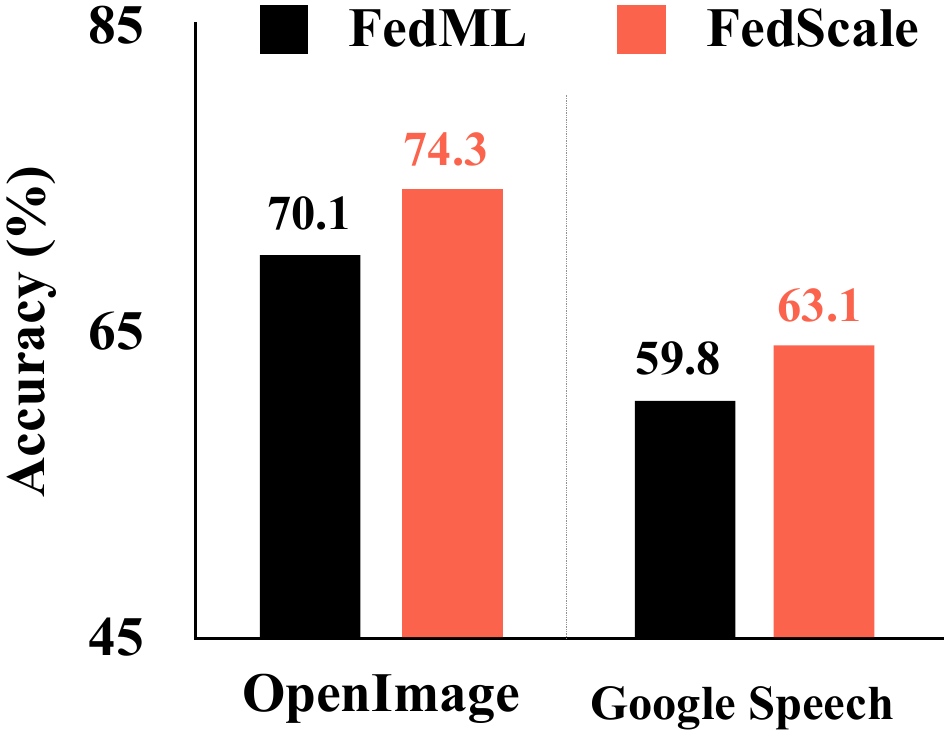}}
  % \end{minipage}
\caption{\name can benchmark the statistical FL performance. (c) shows existing benchmarks can under-report the FedYoGi performance as they cannot support a large number of participants.}
      \label{fig:stats-convergence}
\end{figure*}

\paragraph{\platform is scalable and efficient}
In the simulation mode, \platform can perform large-scale simulations (thousands of clients per round) in both standalone (single CPU/GPU) and distributed (multiple machines) settings. 
This is because: 
\begin{enumerate*}[label=({\arabic*})]
\item \platform uses GPU sharing techniques~\cite{Salus} to divide GPU among tasks so that multiple client simulators can co-locate on the same GPU; 
\item our resource manager monitors the fine-grained resource utilization of machines, queues the overcommitted execution requests, adaptively dispatches requests of the client across machines to achieve load balance, and then orchestrates the simulation based on the client virtual clock.
Instead, state-of-the-art platforms can hardly support the practical FL scale, due to their limited support for distributed evaluations (\eg, FedJax~\cite{fedjax}), and/or the reliance on the traditional ML architecture that trains on a few workers with long-running computation, whereas \platform minimizes the overhead (\eg, frequent data serialization) in the fleet training of FL clients.
As shown in Figure~\ref{fig:scalability}
\footnote{We train ShuffleNet on OpenImage classification task on 10 GPU nodes. Detailed experimental setups in Appendix~\ref{app:eval}.}, other than being able to evaluate the practical FL runtime, \platform not only runs faster than FedML~\cite{fedml} and Flower~\cite{flower}, but it can support large-scale evaluations efficiently. 
\end{enumerate*}

% !TeX root = ../flbench.tex
\section{Experiments}
\label{sec:eval}

In this section, 
we show how \name can facilitate better benchmarking of FL efforts over its counterparts.
% , 
% and highlight important observations to improve practical FL.

\paragraph{Experimental setup}
We use 10 NVIDIA Tesla P100 GPUs in our evaluations. 
Following the real FL deployments~\cite{federated-learning, ggkeyboard}, the aggregator collects updates from the first $N$ completed participants out of $1.3N$ participants to mitigate system stragglers in each round, and $N=100$ by default. 
We experiment with representative \name datasets in different scales and tasks (detailed experiment setup in Appendix~\ref{app:eval}). 

\subsection{How Does \name Help FL Benchmarking?}
\label{eval:bench}

\begin{figure*}[t]
  \centering
      \subfigure[\name reports realistic FL clock. \label{fig:impact-k-sys}]{\includegraphics[width=0.3\linewidth]{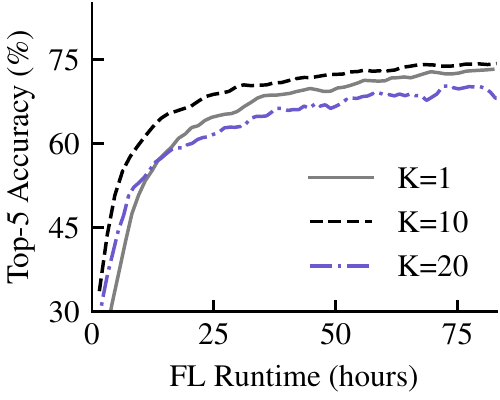}}\hfill
      \subfigure[\name enables fast-forward eval. \label{fig:eval-time}]{\includegraphics[width=0.28\linewidth]{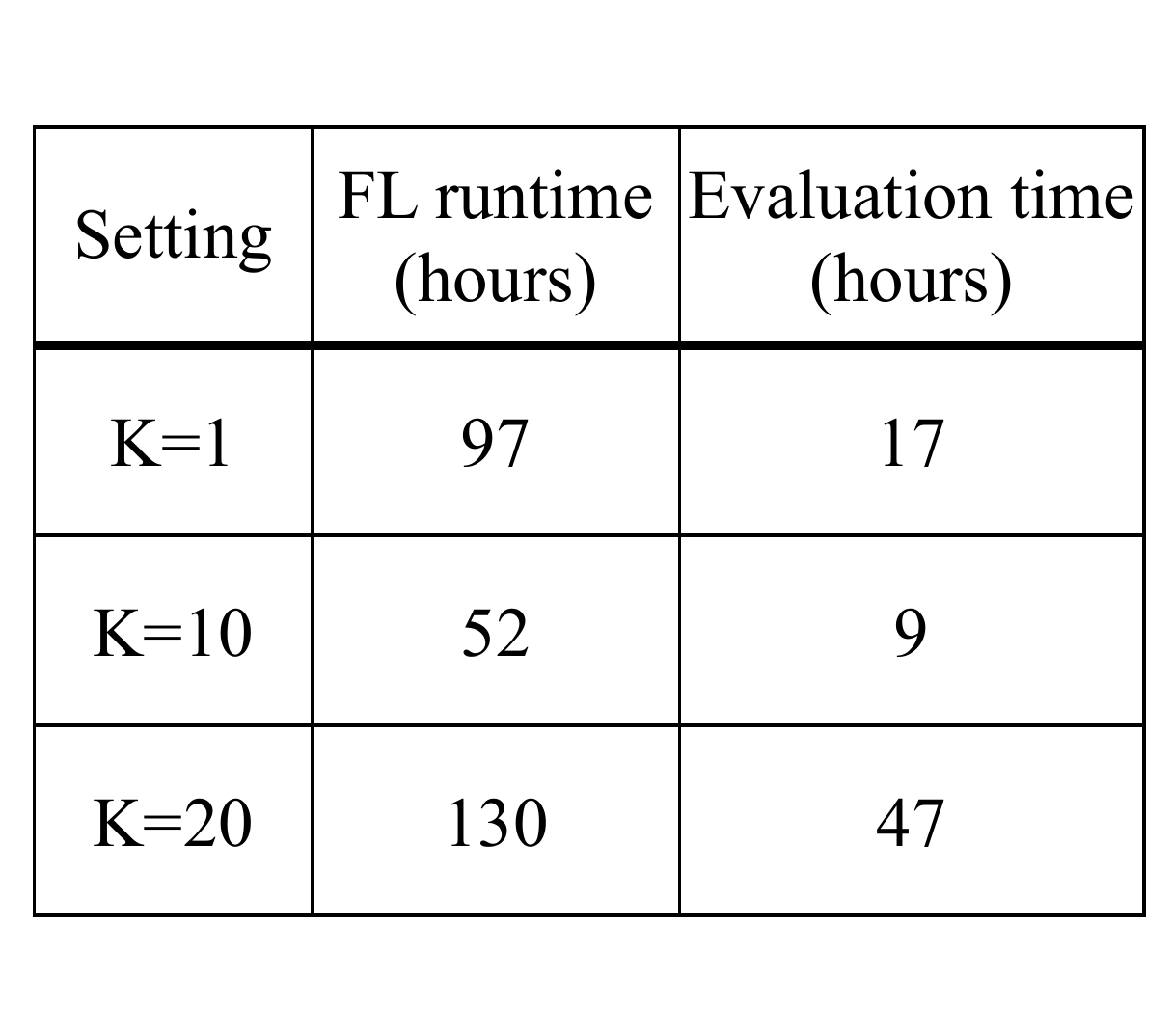}}\hfill
      \subfigure[\name reports FL communication cost. \label{fig:net-cost}]{\includegraphics[width=0.33\linewidth]{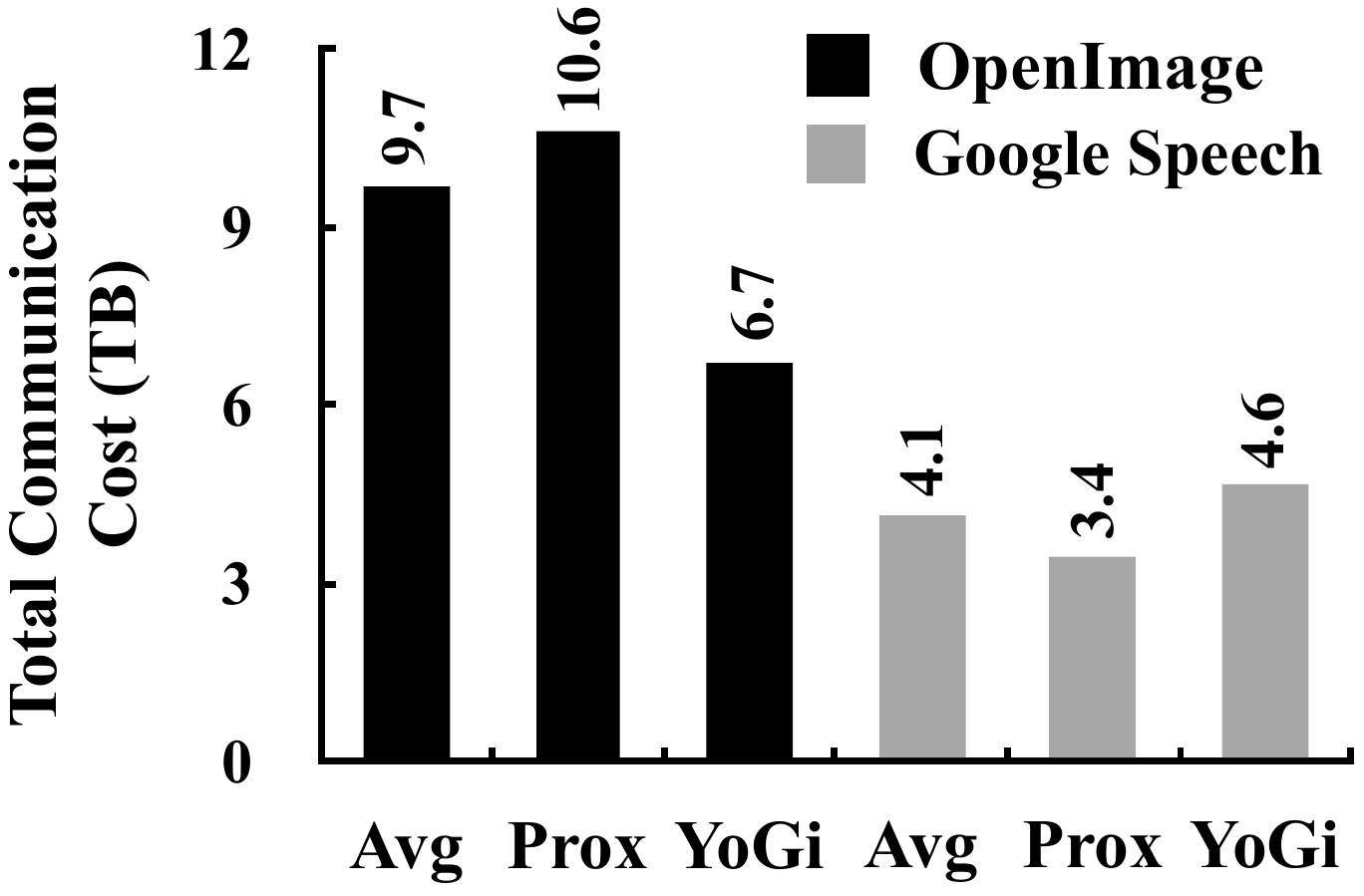}}

      \caption{\name can benchmark realistic FL runtime. (a) and (b) report FedYoGi results on OpenImage with different number of local steps (K); (b) reports the FL runtime to reach convergence.}
      \label{fig:sys-convergence}
      \vspace{-.3cm}
\end{figure*}

Existing benchmarks are insufficient to evaluate the various metrics needed in today's FL.
 % and can even mis-report the performance due to the inability to reproduce the FL setting. 
We note that the performance of existing benchmarks and \name are quite close in the same settings if we turn off the optional system traces in \name. Because the underlying training and FL protocols in evaluations are the same. However, the limited scalability can mislead the practical FL performance. 
Next, we show the effectiveness of \name in benchmarking different FL aspects over its counterparts.

\paragraph{Benchmarking FL statistical efficiency.}
\name provides various realistic client datasets to benchmark the FL statistical efficiency. 
Here, we experiment with state-of-the-art optimizations (FedAvg, FedProx and FedYoGi) -- each aims to mitigate the data heterogeneity -- and the traditional IID data setting. 
Figure~\ref{fig:stats-convergence} and Table~\ref{table:eval} report that: 
(1) the round-to-accuracy performance and final model accuracy of the non-IID setting is worse than that of the IID setting, which is consistent with existing findings~\cite{fl-survey}; 
(2) different tasks can have different preferences on the optimizations. 
For example, FedYoGi performs the best on OpenImage, but it is inferior to FedAvg on Google Speech. 
With much more FL datasets, \name enables extensive studies of the sweet spot of different optimizations; and 
(3) existing benchmarks can under-report the FL performance due to their inability to reproduce the FL setting. 
Figure~\ref{fig:impact-n} reports the final model accuracy using FedML and \name, 
where we attempt to reproduce the scale of practical FL with 100 participants per round in both frameworks, 
but FedML can only support 30 participants because of its suboptimal scalability, which under-reports the FL performance that the algorithm can indeed achieve.

% \paragraph{Why we need to take care of the system efficiency?}
\paragraph{Benchmarking FL system efficiency.}
Existing system optimizations for FL focus on the practical runtime (\eg, wall-clock time in real FL training) 
and the FL execution cost. Unfortunately, existing benchmarks can hardly evaluate the FL runtime due to the lack of realistic system traces, but we now show how \name can help such benchmarking:
\begin{enumerate*}[label=({\arabic*})]
\item \platform enables fast-forward evaluations of the practical FL wall-clock time with fewer evaluation hours. Taking different number of local steps $K$ in local SGD as an example~\cite{fedavg}, Figure~\ref{fig:impact-k-sys} and Table~\ref{fig:eval-time} illustrate that \name can evaluate this impact of $K$ on practical FL runtime in a few hours.  
%reports the FL runtime of ( versus Figure~\ref{fig:impact-k-sys}), while using much fewer evaluation hours (Table~\ref{fig:eval-time}). 
This allows the developer to evaluate large-scale system optimizations efficiently; and 

\item \platform can dictate the FL execution cost by using realistic system traces. 
For example, Figure~\ref{fig:net-cost} reports the practical FL communication cost in achieving the performance of Figure~\ref{fig:stats-convergence}, while Figure~\ref{fig:straggler} reports the system duration of individual clients.   
These system metrics can facilitate developers to navigate the accuracy-cost trade-off. 

\end{enumerate*}

\begin{figure}[t]
\centering
\begin{minipage}{.47\columnwidth}
\centering
  \centering
  \includegraphics[scale=0.75]{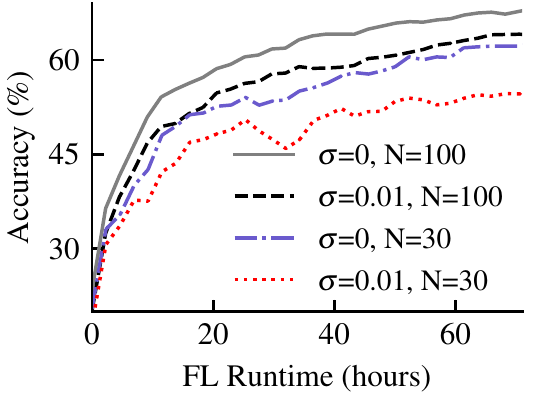}
  \captionof{figure}{\name can benchmark privacy efforts in more realistic FL settings.}
  \label{fig:privacy}
% \caption{Add plugins by inheritance.}
\end{minipage}
\hfill
\begin{minipage}{.47\columnwidth}
\centering
  \centering
  \includegraphics[scale=0.75]{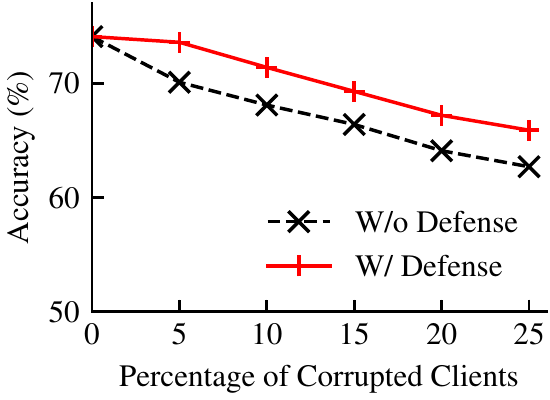}
  \captionof{figure}{\name can benchmark security optimizations with realistic FL data.}
  \label{fig:malicious}
% \caption{Add plugins by inheritance.}
\end{minipage}
\vspace{-.2cm}
\end{figure}

\paragraph{Benchmarking FL privacy and security.} 
\name can evaluate the statistical and system efficiency for privacy and security optimizations more realistically.  
Here, we give an example of benchmarking DP-SGD~\cite{dpfl, dp-sgd}, 
which applies differential privacy to improve the client privacy. 
We experiment with different privacy targets $\sigma$ ($\sigma$=0 indicates no privacy enhancement) and different number of participants per round $N$. 
Figure~\ref{fig:privacy} shows that the scale of participants (\eg, $N$=30) that today's benchmarks can support can mislead the privacy evaluations too: for $\sigma$=0.01, while we notice great performance degradation (12.8\%) in the final model accuracy when $N$=30, this enhancement is viable in practical FL ($N$=100) with decent accuracy drop (4.6\%). Moreover, \name is able to benchmark more practical FL metrics, such as wall-clock time, communication cost added in privacy optimizations, and the number of rounds needed to leak the client privacy under realistic individual client data and Non-IID distributions.

As for benchmarking FL security, we follow the example setting of recent backdoor attacks~\cite{backdoor-clip, backdoor-fl} on the OpenImage, where corrupted clients flip their ground-truth labels to poison the training. 
We benchmarked two settings: one without security enhancement, while the other clips the model updates as~\cite{backdoor-clip}. 
As shown in Figure~\ref{fig:malicious}, while state-of-the-art optimizations report this can mitigate the attacks without
hurting the overall performance on their synthesized datasets, large accuracy drops can occur in more practical FL settings. 

\subsection{Opportunities for Future FL Optimizations}
\label{eval:oppo}

Next, we show \name can shine light on the need for yet unexplored optimizations owing to its realistic FL settings.

\begin{figure}
\centering
\begin{minipage}{.53\columnwidth}
  \centering
  \includegraphics[width=\textwidth]{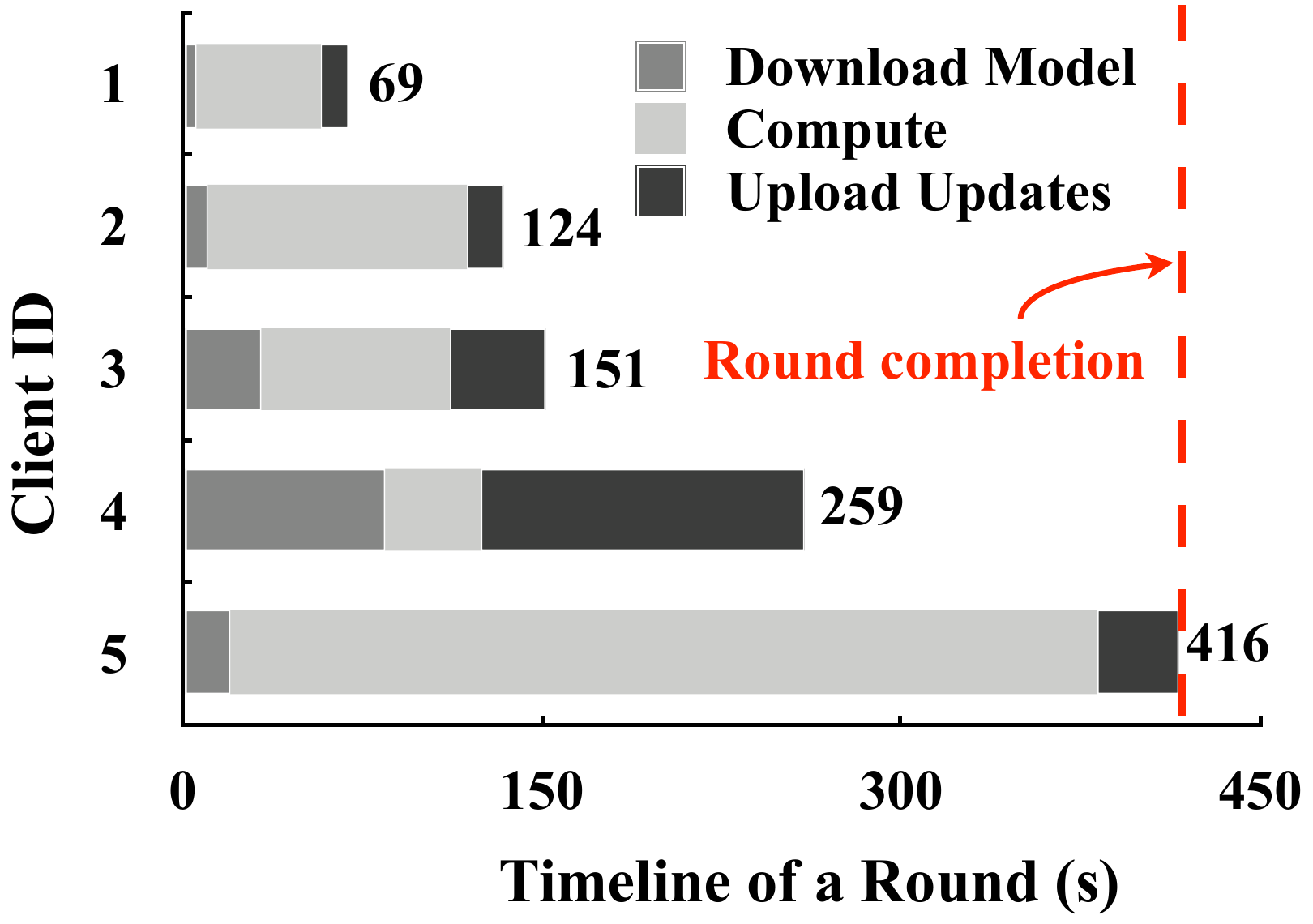}
  \captionof{figure}{System stragglers slow down practical FL greatly.}
  \label{fig:straggler}
\end{minipage}%
\hfill
\begin{minipage}{.43\columnwidth}
\centering
  \centering
  \includegraphics[width=\textwidth]{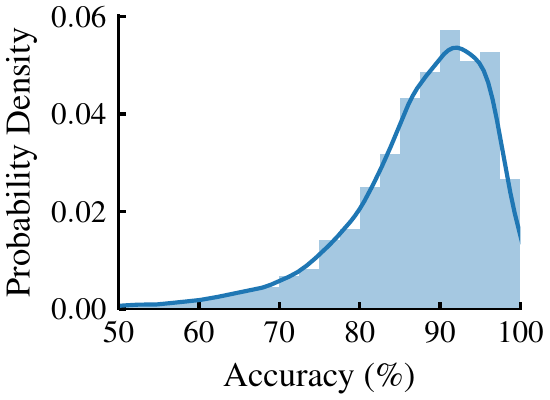}
  \captionof{figure}{Biased accuracy distributions of the trained model across clients (Shufflenet on OpenImage).}
  \label{fig:heter_acc}
% \caption{Add plugins by inheritance.}
\end{minipage}
% \vspace{-.2cm}
\end{figure}

\paragraph{Heterogeneity-aware co-optimizations of communication and computation} 
Existing optimizations for the system efficiency often apply the same strategy on all clients (\eg, using the same number 
 of local steps~\cite{fedavg} or compression threshold~\cite{fetchsgd}), 
while ignoring the heterogeneous client system speed. 
When we outline the timeline of 5 randomly picked 
participants in training of the ShuffleNet (Figure~\ref{fig:straggler}), 
we find: 
(1) system stragglers can greatly slow down the round aggregation in practical FL; and 
(2) simply optimizing the communication or computation efficiency may not lead to faster rounds, 
as the last participant can be bottlenecked by the other resource. 
Here, optimizing the communication can greatly benefit \emph{Client 4}, but it achieves marginal improvement on the round duration as \emph{Client 5} is computation-bound. 
This implies an urgent need of heterogeneity-aware co-optimizations of communication and computation efficiency.

\paragraph{Co-optimizations of statistical and system efficiency} 
Most of today's FL efforts focus on either 
optimizing the statistical or the system efficiency, 
whereas we observe a great opportunity to jointly optimize both efficiencies: 
\begin{enumerate*}[label=({\arabic*})]
\item As the system behavior determines the availability of client data, 
predictable system performance can benefit statistical efficiency. 
For example, in alleviating the biased model accuracy (Figure~\ref{fig:heter_acc}),  
we may prioritize the use of upcoming offline clients to curb the upcoming distribution drift of client data;  
%Moreover, the popular overcommittment in selecting clients can deemphasize stragglers, leading to poor accuracy on slow clients; and 
\item Statistical optimizations should be aware of the heterogeneous client system speed. 
For example, instead of applying one-fit-all strategies (\eg, local steps or gradient compression) for all clients, 
%faster workers can trade more system latency against better statistical benefits. For example, 
faster workers can trade more system latency for more statistical benefits (\eg, transferring more traffics with less intensive compression).
%we may enforce less intensive gradient compression on faster workers for more accurate updates.  
\end{enumerate*}

\paragraph{FL design-decisions considering mobile environment} 
Existing efforts have largely overlooked the interplay of client devices and training speed (\eg, using a large local steps to save communication~\cite{fedavg}), 
%While recent FL efforts reports that using larger local steps can greatly save communication cost~\cite{fedavg}, 
however, as shown in Figure~\ref{fig:mobile}, running intensive on-device computation for a long time can quickly drain the battery, or even burn the device, leading to the unavailability of clients. Therefore, we believe that a power and temperature-aware training algorithm (\eg, different local steps across clients or device-aware NAS) can be an important open problem.

\section{Conclusion}
\label{sec:conclusion}

To enable scalable and reproducible FL research, we introduce \name, 
a diverse set of realistic FL datasets in terms of scales, task categories, and client system behaviors, 
along with a scalable and extensible evaluation platform, \platform.
% We provide realistic federated datasets for benchmarking today's FL efforts. 
%To enable efficient and standardized FL evaluations, 
%We also introduce, \platform, a more scalable evaluation platform than the existing. 
\platform performs fast-forward evaluation of the practical FL runtime metrics needed in today's works. 
More importantly, \platform provides ready-to-use realistic datasets and flexible APIs to allow more FL applications, such as benchmarking NAS, model inference, and a broader view of federated computation (\eg, multi-party computation). 
\name is open-source at \url{http://fedscale.ai}, and we hereby invite the community to develop and contribute
state-of-the-art FL efforts. 
% \paragraph{Societal Impacts and Limitations}
% We expect \name to be a standard benchmark in federated learning, contributing to the significant advancements of the field. 
% One potential negative impact is that FedScale might narrow down the scope of future papers to the tasks and dataset types that have been included so far. In order to mitigate such a negative impact and limitation, 
% we have made \name open-source at: \url{https://github.com/SymbioticLab/FedScale}, and will regularly update
% our datasets and tasks, based on the input from the community.

% \mosharaf{Now that I think about it, we should at least leave a pointer to tuning, NAS, and inference benchmarks as future work in the end.}

\section*{Acknowledgments}
\label{sec:ack}

We would like to thank the anonymous reviewers and SymbioticLab members for their insightful feedback. 
We also thank \name contributors and users from many different academic institutions and industry for their valuable inputs. 
This work was supported in part by NSF grants CNS-1900665, CNS-1909067, and CNS-2106184, and a grant from Cisco.

\bibliographystyle{icml2022}
\bibliography{flbench}
\clearpage
\newpage
% !TeX root = ../kuiper-osdi2021.tex
\appendix

\setlength{\tabcolsep}{3pt}
\begin{figure*}[t]
% \small
\begin{minipage}{\textwidth}
\small
\begin{center}
\begin{tabular}{cccrrc}
\toprule  
Category		& Name 		&Data Type 	& \#Clients	& \#Instances		& Example Task \\
\midrule  

\multirow{7}{*}{\textbf{CV}} 
 	& \names{iNature}~\cite{inatural}	&Image	&	2,295	&	193K		&	Classification	\\
 	& \names{FEMNIST} \cite{FEMNIST}			&Image	&	3,400	&	640K		&	Classification		\\
  	& \names{OpenImage}~\cite{openimg}			&Image	&	13,771	&	1.3M		&	Classification, Object detection		\\	
  	& \names{Google Landmark}~\cite{landmark}	&Image	&	43,484	&	3.6M		&	Classification		\\
  	& \names{Charades}	\cite{charades}		&Video	&	266		&	10K			&	Action recognition		\\
	& \names{VLOG}	\cite{vlog}	&Video	&	4,900		&	9.6K	    &	Classification, Object detection		\\
	& \names{Waymo Motion}~\cite{waymo-motion}	&Video	&	496,358		&	32.5M			&	Motion prediction		\\	[.5ex]
\hline \\[-2.ex]
\multirow{9}{*}{\textbf{NLP}} 
	& \names{Europarl}~\cite{europarl}	&Text	& 27,835		&	1.2M		&	Text translation	\\
	& \names{Blog Corpus}~\cite{blog-corpus}	&Text	&   19,320		&	137M		&	Word prediction		\\
	& \names{Stackoverflow}~\cite{stack-overflow}	&Text	&  342,477		&	135M		&	Word prediction, Classification	\\
	& \names{Reddit}~\cite{reddit}		&Text	&  1,660,820	&	351M		&	Word prediction	\\
	& \names{Amazon Review}	\cite{amazon} &Text	&  1,822,925	&	166M		&	Classification, Word prediction	\\
	& \names{CoQA}	\cite{coqa}			&Text 	& 7,685			&	116K		& 	Question Answering \\
	& \names{LibriTTS}	\cite{LibriTTS}	&Text	& 2,456			&	37K			&	Text to speech \\
	& \names{Google Speech}~\cite{google-speech} &Audio	& 2,618			&	105K		&	Speech recognition	\\
	& \names{Common Voice}~\cite{common-voice}	&Audio	& 12,976		&	1.1M		&	Speech recognition \\[.5ex]
\hline \\[-2.ex]
\multirow{4}{*}{\textbf{Misc ML}} 
& \names{Taxi Trajectory}	&Text   &442		&  1.7M		& Sequence prediction \\
	& \names{Taobao}~\cite{taobao}	&Text	& 182,806		&	20.9M		&	Recommendation	\\
	& \names{Puffer Streaming}~\cite{puffer}	&Text   &121,551		&  15.4M		& Sequence prediction \\
	& \names{Fox Go}~\cite{go-data}	&Text	& 150,333		&	4.9M		&	Reinforcement learning	\\
\bottomrule 
\end{tabular}
\end{center}
\captionof{table}{Statistics of \name datasets. \name has 20 realistic client datasets, which are from the real-world collection, and we partitioned each dataset using its real client-data mapping.}
\vspace{-.2cm}
\label{table:full-data}
\end{minipage}
\end{figure*}

\section{Experiment Setup}
\label{app:eval}

\paragraph{Scalability Evaluations}
We evaluate the scalability of \platform, FedML (GitHub \textit{commit@2ee0517}) and Flower (v0.17.0 atop Ray v1.9.2) using a cluster with 10 GPU nodes. Each GPU node has a P100 GPU with 12GB GPU memory and 192GB RAM. We train the ShuffleNet-V2 model on the OpenImage dataset. We set the minibatch size of each participant to 32, and the number of local steps $K$ to 20, which takes around 2800MB GPU memory for each model training. As such, we allow each GPU node to run 4 processes in benchmarking these three frameworks. 

\paragraph{Evaluation Setup}
Applications and models used in our evaluations are widely used on mobile devices. 
We set the minibatch size of each participant to 32, and the number of local steps $K$ to 20. 
We cherry-pick the hyper-parameters with grid search, ending up with an initial learning rate 0.04 for CV tasks and 5e-5 for NLP tasks. The learning rate decays by 0.98 every 10 training rounds. These settings are consistent with the literature. More details about the input dataset are available in Appendix~\ref{app:dataset}.

\section{Introduction of \name Datasets}
\label{app:dataset}

\name currently has 20 realistic federated datasets across a wide range of scales and task categories (Table~\ref{table:full-data}). Here, we provide the description of some representative datasets.
%, and the reader can refer to \name repository (\url{https://github.com/SymbioticLab/flperf}) for more datasets. 

\paragraph{Google Speech Commands.} 
A speech recognition dataset \cite{google-speech} with over ten thousand clips of one-second-long duration. Each clip contains one of the 35 common words (\eg, digits zero to nine, "Yes", "No", "Up", "Down") spoken by thousands of different people. 

\paragraph{OpenImage.} 
OpenImage \cite{openimg} is a vision dataset collected from Flickr, an image and video hosting service. 
It contains a total of 16M bounding boxes for 600 object classes (\eg, Microwave oven). We clean up the dataset according to the provided indices of clients.

\paragraph{Reddit and StackOverflow.} 
Reddit \cite{reddit} (StackOverflow \cite{stack-overflow}) consists of comments from the Reddit (StackOverflow) website. It has been widely used for language modeling tasks, and we consider each user as a client. In this dataset, we restrict to the 30k most frequently used words, and represent each sentence as a sequence of indices corresponding to these 30k frequently used words. 

\paragraph{VLOG.} VLOG \cite{vlog} is a video dataset collected from YouTube. It contains more than 10k Lifestyle Vlogs, videos that people purportedly record to show their lives, 
from more than 4k actors. This dataset is aimed at understanding everyday human interaction and contains labels for scene classification, hand-state prediction, and hand detection tasks.

\paragraph{LibriTTS.} LibriTTS \cite{LibriTTS} is a large-scale text-to-speech dataset. It is derived from audiobooks in  LibriVox project. There are 585 hours of read English speech from 2456 speakers at a 24kHz sampling rate. 

\paragraph{Taobao.} Taobao Dataset \cite{taobao} is a dataset of click rate prediction about display Ad, which is displayed on the website of Taobao. It is composed of 1,140,000 users ad display/click logs for 8 days, which are randomly sampled from the website of Taobao. 
%We partitioned it using its real client-data mapping.

\paragraph{Waymo Motion.} Waymo Motion \cite{waymo-motion} is composed of 103,354 segments each containing 20 seconds of object tracks at 10Hz and map data for the area covered by the segment. These segments are further broken into 9 second scenarios, and we consider each scenario as a client. 

\paragraph{Puffer Streaming.} Puffer is a Stanford University research study about using machine learning (\eg, reinforcement learning~\cite{pensieve}) to improve video-streaming algorithms: the kind of algorithms used by services such as YouTube, Netflix, and Twitch. Puffer dataset~\cite{puffer} consists of 15.4M sequences of network throughput on edge clients over time.  

% We will make \name open-source on the Github. 
% So the code and dataset can be downloaded from the repository. 
% For each dataset, we will provide detailed descriptions (\texttt{README.md}) of the source, organization, format and use case under the repository. 
% In the future, these datasets will be hosted on our server, or be migrated to the stable storage of AWS. 
% For the evaluation platform \platform, we will provide the configuration and job submission guidelines as well. 

\section{Comparison with Existing FL Benchmarks}
\label{app:comparison}

In this section, we compare \name with existing FL benchmarks in more details.

\paragraph{Data Heterogeneity} 
Existing benchmarks for FL are mostly limited in the variety of realistic datasets for real-world FL applications. 
Even they have various datasets (\eg, LEAF~\cite{leaf-bench}) and FedEval~\cite{fedeval}), their datasets are mostly synthetically derived from conventional datasets (\eg, CIFAR) and limited to quite a few FL tasks.
These statistical client datasets can not represent realistic characteristics and are inefficient to benchmark various real FL applications.  
Instead, \name provides 20 comprehensive realistic datasets for a wide variety of tasks and across small, medium, and large scales, which can also be used in data analysis to motivate more FL designs.

\paragraph{System Heterogeneity} 
The practical FL statistical performance also depends on the system heterogeneity (\eg, internet bandwidth~\cite{relay-hotcloud} and client availability), which has inspired lots of optimizations for FL system efficiency.
However, existing FL benchmarks have largely overlooked the system behaviors of FL clients, which can produce misleading evaluations, and discourage the benchmarking of system efforts. 
To emulate the heterogeneous system behaviors in practical FL, 
\name incorporates real-world traces of mobile devices, associates each client with his system speeds, as well as the availability. Moreover, we develop a more efficient evaluation platform, \platform, to automate FL benchmarking.

\paragraph{Scalability} 
Existing frameworks, perhaps due to the heavy burden of building complicated system support, largely rely on the traditional ML architectures (e.g., the primitive parameter-server architecture of Pytorch). These architectures are primarily designed for the traditional large-batch training on a number of workers, and each worker often trains a single batch at a time. However, this is ill-suited to the simulation of thousands of clients concurrently: (1) they lack tailored system implementations to orchestrate the synchronization and resource scheduling, for which they can easily run into synchronization/memory issues and crash down; (2) their resource can be under-utilized, as FL evaluations often use a much smaller batch size.

Tackling all these inefficiencies requires domain-specific system designs.
%, and the \platform is refactored atop of our Oort project~\cite{kuiper-osdi}. 
Specifically, we built a cluster resource scheduler that monitors the fine-grained resource utilization of machines, queues the overcommitted simulation requests, adaptively dispatches simulation requests of the client across machines to achieve load balance, and then orchestrates the simulation based on the client virtual clock. Moreover, given a much smaller batch size in FL, we maximize the resource utilization by overlapping the communication and computation phrases of different client simulations. The former and the latter make \name more scalable across machines and on single machines, respectively.

Empirically, we have run the 20-GPU set up on different datasets and models in Figure~\ref{fig:scalability} and Table~\ref{table:20-tpt}, and are aware of at least one group who ran \platform with more than 60 GPUs~\cite{kuiper-osdi}. 

% \vspace{-.3cm}
\renewcommand{\arraystretch}{1.1}
\setlength{\tabcolsep}{6pt}
\begin{table}[h]
\centering % used for centering table
\footnotesize
\setlength{\arrayrulewidth}{0.9pt}
\begin{tabularx}{\linewidth}{c | c | c | c |c} % centered columns (4 columns)
\hhline{--|-|-|-} 
								 &	\multicolumn{4}{c}{Eval. Duration/Round vs. \# of Clients/Round} \\
\hhline{~-|-|-|-} 
								&  10  & 100  & 1K  & 10K  \\
\hhline{--|-|-|-} 
\textbf{\name}        &		\textbf{0.03 min}			&  	\textbf{0.16 min}       &  \textbf{1.14 min}     &  \textbf{10.9 min}  \\
\hhline{-|-|-|-|-} 
FedML        &		0.58 min			&  	4.4 min       &  fail to run     &  fail to run  \\
\hhline{-|-|-|-|-} 
Flower        &		0.05 min			&  	0.23 min       &  2.4 min     &  fail to run  \\
\hhline{-|-|-|-|-} 
\end{tabularx}
\caption{\name is more scalable and faster. Image classification on iNature dataset using MobileNet-V2 on 20-GPU setting.} 
\label{table:20-tpt}
%\vspace{-.48cm}
\end{table}
\vspace{-.3cm}

\begin{figure}[h]
\centering
\begin{lstlisting}
from fedscale.core.client_manager import ClientManager
import Oort

class Customized_ClientManager(ClientManager):
	def __init__(self, *args):
		super().__init__(*args)
		self.oort_selector = Oort.create_training_selector(*args)

  # Replace default client selection algorithm w/ Oort
  def select_participants(self, numOfClients, cur_time, feedbacks):
		# Feed Oort w/ execution feedbacks
		oort_selector.update_client_info(feedbacks)
		selected_clients = oort_selector.select_participants(
			numOfClients, cur_time)

		return selected_clients
\end{lstlisting}
\captionof{figure}{Evaluate client selection algorithm~\cite{kuiper-osdi}.}
% \caption{Add plugins by inheritance.}
\label{fig:selection}
\vspace{-.3cm}
\end{figure}%

\begin{figure}[h]
\begin{minipage}{.498\textwidth}
\centering
\begin{lstlisting}
from fedscale.core.client import Client

class Customized_Client(Client):
# Customize the training on each client
  def train(self,client_data,model,conf):
  		# Get the training result from 
  		# the default training component
		training_result = super().train(
				client_data, model, conf)

		# Implementation of compression
		compressed_result = compress_impl(training_result)
		return compressed_result
\end{lstlisting}
\captionof{figure}{Evaluate model compression~\cite{fetchsgd}.}
% \caption{Add plugins by inheritance.}
\label{fig:compression}
\end{minipage}
\hfill 
\begin{minipage}{.498\textwidth}
\centering
\begin{lstlisting}
from fedscale.core.client import Client

class Customized_Client(Client):
# Customize the training on each client
  def train(self,client_data,model,conf):
		training_result = super().train(
				client_data, model, conf)

		# Clip updates and add noise
		secure_result = secure_impl(training_result)
		return secure_result
\end{lstlisting}
\captionof{figure}{Evaluate security enhancement~\cite{backdoor-clip}.}
% \caption{Add plugins by inheritance.}
\label{fig:secure-backdoor}
\end{minipage}
\vspace{-.3cm}
\end{figure}

\begin{figure}[h]
	\begin{minipage}{.498\textwidth}
	\centering
	\begin{lstlisting}[escapechar=@]
		import flwr as fl
		
		def get_config_fn():
			# Implementation of randomly selection
			client_ids = @\mytikzmark{h1start}@random_selection@\mytikzmark{h1end}@()
			config = {"ids": client_ids}
			return config

		# Customized Strategy
		strategy = @\mytikzmark{h2start}@CustomizedStrategy@\mytikzmark{h2end}@(
			on_fit_config_fn=get_config_fn())

		fl.server.start_server(
			config={"num_rounds":args.round},
			strategy=strategy)
	\end{lstlisting}
	\begin{tikzpicture}[remember picture, overlay]
		\highlight{h1start}{h1end}
		\highlight{h2start}{h2end}
	\end{tikzpicture}
	% \caption{Add plugins by inheritance.}
	
	\end{minipage}
	\hfill 
	\begin{minipage}{.498\textwidth}
	\centering
	\begin{lstlisting}[escapechar=@]
	import flwr as fl

	class Customized_Client():
		def fit(self, config, net):
			# Customization of client data
			trainloader = @\mytikzmark{h1start}@select_dataset@\mytikzmark{h1end}@(
				config["ids"][args.partition])
			train(net, trainloader)
			compressed_result = self.get_parameters()
			# Implementation of compression
			compressed_result = compress_impl(
					training_result)
			return compressed_result
	
	fl.client.start_numpy_client(
		args.address, client=CustomizedClient())
	\end{lstlisting}
	\begin{tikzpicture}[remember picture, overlay]
		\highlight{h1start}{h1end}
	\end{tikzpicture}
	% \caption{Add plugins by inheritance.}
	\end{minipage}
	\captionof{figure}{Evaluate model compression with Flower~\cite{flower}. The developer needs to implement the functions in grey by his own. Note that each function can take tens of lines of code.}
	\label{fig:flower-compression} 
	\end{figure}

\paragraph{Modularity} 
As shown in Table~\ref{table:comparison}, some existing frameworks (e.g., LEAF and FedEval) do not provide user-friendly modularity, which requires great engineering efforts to benchmark different components, and we recognize that FedML and Flower provide the API modularity too.

On the other hand, \platform's modularity for easy deployments and broader use cases is not limited to APIs (Figure~\ref{fig:arc}): (1) \platform Data Loader: it simplifies and expands the use of realistic datasets. \eg, developers can load and analyze the realistic FL data to motivate new algorithm designs, or imports new datasets/customize data distributions in \name evaluations; (2) Client simulator: it emulates the system behaviors of FL clients, and developers can customize their system traces in evaluating the FL system efficiency too; (3) Resource manager: it hides the system complexity in training large-scale participants. 

\section{Examples of New Plugins}
\label{app:plugin}

In this section, we demonstrate examples to show the ease of integrating today's FL efforts in \name evaluations. 

At its core, \platform provides flexible APIs on each module so that the developer can access and customize methods of the base class. Table~\ref{table:apis} illustrates some example APIs that can facilitate diverse FL efforts. Note that \platform will automatically integrate new plugins into evaluations, and then produces practical FL metrics.  
Figure~\ref{fig:selection} demonstrates that we can easily evaluate new client selection algorithms, Oort~\cite{kuiper-osdi}, by modifying a few lines of the \texttt{clientManager} module. Similarly, Figure~\ref{fig:compression} and Figure~\ref{fig:secure-backdoor} show that we can extend the basic \texttt{Client} module to apply new gradient compression~\cite{fetchsgd} and enhancement for malicious attack~\cite{backdoor-clip}, respectively.  

\paragraph{Comparison with other work} 
Figure~\ref{fig:flower-compression} shows the same evaluation of gradient compression~\cite{fetchsgd} as that in Figure~\ref{fig:compression} using flower~\cite{flower}, which requires much more human efforts than Figure~\ref{fig:compression}. The gray components in figure~\ref{fig:flower-compression} requires implementation. Flower falls short in providing APIs for passing metadata between client and server, for example client id, which makes server and running workers client-agnostic. To customized anything for clients during FL training, developers have to go through the source code and override many components to share client meta data between server and workers.

\end{document}